\documentclass[conference]{IEEEtran}
\IEEEoverridecommandlockouts
\usepackage{cite}
\usepackage{amsmath,amssymb,amsfonts}
\usepackage{algorithmic}
\usepackage{graphicx}
\usepackage{textcomp}
\usepackage{xcolor}
\usepackage{microtype}
\usepackage[justification=centering]{caption}

\usepackage[noend,linesnumbered,ruled]{algorithm2e}

\usepackage{todonotes}
\usepackage{multirow}
\usepackage{subcaption}

\newcommand{\cmaes}{CMAES}

\newcommand{\canonical}{Canonical ES}

\newcommand{\gdr}{GDR}
\newcommand{\longgdr}{Genetic Drift Regularization}

\newcommand{\sgdr}{GDR²}
\newcommand{\longsgdr}{Squared Genetic Drift Regularization}

\newcommand{\mujoco}{\textsc{MuJoCo}}

\newcommand{\halfcheetah}{\textsc{HalfCheetah}}
\newcommand{\walker}{\textsc{Walker2D}}
\newcommand{\hopper}{\textsc{Hopper}}
\newcommand{\humanoid}{\textsc{Humanoid}}
\newcommand{\swimmer}{\textsc{Swimmer}}
\newcommand{\ant}{\textsc{Ant}}

\begin{document}

\title{\longgdr{}: on preventing Actor Injection from breaking Evolution Strategies\\}

\author{
\IEEEauthorblockN{Paul Templier}
\IEEEauthorblockA{
    \textit{ISAE-SUPAERO,}\\
    \textit{Université de Toulouse,}\\
Toulouse, France \\
paul.templier@isae.fr}

\and

\IEEEauthorblockN{Emmanuel Rachelson}
\IEEEauthorblockA{
    \textit{ISAE-SUPAERO,}\\
    \textit{Université de Toulouse,}\\
Toulouse, France \\
emmanuel.rachelson@isae.fr}

\and

\IEEEauthorblockN{Antoine Cully}
\IEEEauthorblockA{
    \textit{Imperial College London,}\\
London, United Kingdom \\
a.cully@imperial.ac.uk}
\and

\IEEEauthorblockN{Dennis G. Wilson}
\IEEEauthorblockA{
    \textit{ISAE-SUPAERO,}\\
    \textit{Université de Toulouse,}\\
Toulouse, France \\
dennis.wilson@isae.fr}
}

\maketitle

\begin{abstract}

    Evolutionary Algorithms (EA) have been successfully used for the optimization of neural networks for policy search, but they still remain sample inefficient and underperforming in some cases compared to gradient-based reinforcement learning (RL). Various methods combine the two approaches, many of them training a RL algorithm on data from EA evaluations and injecting the RL actor into the EA population. However, when using Evolution Strategies (ES) as the EA, the RL actor can drift genetically far from the the ES distribution and injection can cause a collapse of the ES performance. Here, we highlight the phenomenon of \textit{genetic drift} where the actor genome and the ES population distribution progressively drift apart, leading to injection having a negative impact on the ES. We introduce \longgdr{} (\gdr{})\footnote{https://anonymous.4open.science/r/GDR-E784}, a simple regularization method in the actor training loss that prevents the actor genome from drifting away from the ES. We show that \gdr{} can improve ES convergence on problems where RL learns well, but also helps RL training on other tasks, , fixes the injection issues better than previous controlled injection methods
\end{abstract}


\section{Introduction}
\label{sec:intro}

Gradient-based and gradient-free methods have been used to optimize policy networks for control problems. Reinforcement Learning (RL) algorithms, specifically deep RL, use gradient estimates from individual transitions to maximize expected reward, backpropagating this estimate through a neural network. Gradient-free optimization, mostly in the form of evolutionary algorithms, instead explore the parameter space of policy networks to maximize the sum of reward over entire episodes \cite{chrabaszczBackBasicsBenchmarking2018,salimansEvolutionStrategiesScalable2017}. These two approaches present different advantages, with deep RL algorithms being more sample efficient than evolutionary policy search, and evolutionary algorithms exploring the policy space naturally.

Recently, a number of approaches have been proposed to combine these two approaches \cite{sigaudCombiningEvolutionDeep2022}. In many of these algorithms, which are further detailed below, a gradient-based method is used to optimize a policy in parallel to a gradient-free method, with some information exchange between the two methods. As the gradient-free evolutionary approaches use a population of multiple policies, it is simple to import the gradient-trained RL policy into the evolutionary population. Multiple methods employ this technique \cite{bodnarProximalDistilledEvolutionary2020,khadkaEvolutionGuidedPolicyGradient2018,suriMaximumMutationReinforcement2021}, which is referred to as ``actor injection''. However, in distribution-based methods such as evolutionary strategies (ES), we observe that RL actor parameters drift away from the ES distribution, which can make injecting them detrimental to the ES.

In this work, we first study how actor injection impacts the distance in parameter space between the RL actor and the ES distribution. We introduce \longgdr{}, a method to keep the actor parameters (considered its genome in evolution) close to the ES distribution through a simple regularization term in the actor training loss. Finally, we show that \gdr{} mitigates the impact of actor injection while still allowing for improvement over evolution.

\subsection{Evolution Strategies for Policy Search}
\label{sec:related}

Neural networks with a pre-determined architecture can be optimized as black-box problems with evolutionary methods such as genetic algorithms \cite{mitchellIntroductionGeneticAlgorithms1998} or evolution strategies (ES, \cite{rechenberg1978evolutionsstrategien}) by concatenating the weighs of the network into a vector that is used as a genome. ES are a class of evolutionary algorithms that uses a multivariate Gaussian distribution to sample the next generation of genomes, evaluates them, and updates the distribution parameters to maximize the fitness of the population. Evolution algorithms have a long history of use in policy search, and ES have been recently shown to be competitive with gradient-based RL methods \cite{chrabaszczBackBasicsBenchmarking2018,salimansEvolutionStrategiesScalable2017}.

Canonical ES \cite{chrabaszczBackBasicsBenchmarking2018} is a simple ES that has been shown to perform well on complex control tasks such as Atari \cite{bellemareArcadeLearningEnvironment2012} and the \mujoco{} \cite{todorov2012mujoco} robotics control suite, despite being very simple. It has no step-size adaptation nor covariance matrix adaptation, contrary to \cmaes{} \cite{hansenCompletelyDerandomizedSelfAdaptation2001} or the OpenAI ES \cite{salimansEvolutionStrategiesScalable2017}, and only updates the mean of the distribution from the sampled population and their corresponding fitness values. We study this ES due to its simplicity, especially in its lack of state such as the covariance matrix of \cmaes{} or momentum from OpenAI ES; rather, an independent gradient estimate is determined each step from the most recent population.%

\subsection{Evolutionary Reinforcement Learning}

Multiple methods have proposed combining the gradient-free optimization of evolutionary algorithms with gradient-based Reinforcement Learning \cite{khadkaEvolutionGuidedPolicyGradient2018,marchesiniGENETICSOFTUPDATES2021,pourchotCEMRLCombiningEvolutionary2019,suriMaximumMutationReinforcement2021}; we refer to \cite{sigaudCombiningEvolutionDeep2022} for a comprehensive survey. In \cite{sigaudCombiningEvolutionDeep2022}, two families of methods are identified: ones which use evolution directly to maximize reward (exploitation), and other approaches which use evolution to find a diversity of policies (exploration). We here focus on methods that use evolution and RL to directly maximize reward.

Many of the existing algorithms rely on RL actor injection, where an agent trained with RL in parallel to an evolutionary algorithm is periodically added to the population. This is the case in the Evolutionary Reinforcement Learning (ERL, \cite{khadkaEvolutionGuidedPolicyGradient2018}) and Evolution-based Soft Actor-Critic (ESAC, \cite{suriMaximumMutationReinforcement2021}) algorithms. In these methods, if the policy network parameters (genes) introduced by the RL agent are deemed beneficial, they will be kept in the next generation of the evolutionary algorithm, and if they are not they will be removed. We further discuss actor injection below.

Other methods such as CEM-ACER \cite{tangGuidingEvolutionaryStrategies2021} instead use critic gradient addition, where a RL loop computes a gradient to add to certain individuals in the population of the evolution. If the gradient steps help improve the fitness, the solution will impact the update; otherwise, the updated agents will be eliminated. In CEM-RL \cite{pourchotCEMRLCombiningEvolutionary2019}, the RL algorithm only trains the critic, and the agents from the population are used as actors. Gradients can also be used to inform the ES search, as used on Guided ES \cite{maheswaranathanGuidedEvolutionaryStrategies2019} and Self-Guided ES \cite{liuSelfGuidedEvolutionStrategies2020}. Computing the gradient allows to shape the ES distribution in its direction, hence favoring the search in that direction.

While the principal driver of search in the above methods is the evolutionary algorithm, the RL algorithm can also be used to guide search such as in Soft Updates for Policy Evolution (Supe-RL, \cite{marchesiniGENETICSOFTUPDATES2021}), where evolution is used to periodically update the parameters of the RL agent. During the RL training loop, the methods periodically adds generations of the evolutionary algorithm started from the current RL agent, to help exploration. %
Finally, SC-RL \cite{wangSurrogateAssistedControllerExpensive2022} uses a RL critic to evaluate agents optimized with evolution to reduce evaluation costs, similarly to a surrogate model.

These methods show many ways of combining the strengths of evolution and RL to develop more efficient learning algorithms. There are other combined approaches, such as \cite{nilssonPolicyGradientAssisted2021a}, which combines policy gradient methods with a behavioral diversity exploration algorithm, MAP-Elites. In this work, we focus on the mechanics behind combining these two approaches, which could be used in a variety of algorithms.

As a principle component of mixing EA and RL in the same method is often to reuse samples from evaluations of diverse policies in the EA, off-policy RL algorithms like TD3 are the most used in these methods. These methods, as opposed to on-policy RL algorithms, can benefit from samples drawn by the evaluation of the evolutionary population. Twin Delayed DDPG (TD3, \cite{fujimotoAddressingFunctionApproximation2018}) is an off-policy Value-Iteration RL algorithm for continuous actions control tasks that builds on Deep Deterministic Policy Gradient (DDPG, \cite{lillicrapContinuousControlDeep2015}). DDPG trains a critic network on a replay buffer of transitions to estimate Q-values of state-action pairs, and an actor network to pick actions that maximize its expected Q-value based on the critic's evaluation. The replay buffer is filled by transitions from rollouts of the actor policy in the environment with exploration noise. TD3 improves on DDPG by doubling the critic network. Soft Actor-Critic (SAC, \cite{haarnojaSoftActorCriticOffPolicy2018}) has also been used with ES in ESAC \cite{suriMaximumMutationReinforcement2021} and in \cite{limUnderstandingSynergiesQualityDiversity2023} with a diversity-focused evolutionary algorithm. TD3 has the advantage of using deterministic policies, as used in EA, while other methods like SAC use stochastic policies. We hence use TD3 in the rest of this paper, but the GDR framework could be extended to other off-policy RL methods with an actor like SAC.

Finally, this work offers new insight into the differences between neural networks optimized using backpropagation and using evolutionary strategies. While both approaches have been used for neural network parameter optimization, there is little study on the resultant differences. \cite{langeLotteryTicketsEvolutionary2023} studies the existence of lottery tickets in evolutionary optimization, comparing them to lottery tickets found by backpropagation. In this work, we study how backpropagation and EAs lead to different neural network parameters, even when estimating similar gradients.

\subsection{Actor injection}

Actor injection uses the actor of a RL method trained on the transitions from the EA agents evaluation and injects its parameters as the genome of a new individual into the population of the EA, usually at each generation. The injected actor is evaluated like the other agents, then used in the EA update (Algorithm \ref{alg:actorinjection}).

In an ES, we sample $N-1$ agents from the distribution for a target population size of $N$, where the actor is added as the $Nth$ agent. It is the evaluated and used in the update of the ES distribution.

\begin{algorithm}[h]
    \caption{Actor Injection}
    \label{alg:actorinjection}
    \begin{algorithmic}[1]

        \STATE Initialize environment $\mathcal{E}$
        \STATE Initialize replay buffer $R$
        \STATE Initialize actor $\theta_a$ and critic $\phi$
        \STATE Initialize ES distribution $D$
        \STATE Initialize population size $\lambda$ %

        \FOR{generation $g$ in $G$}
        \STATE Sample $\lambda - 1$ individuals from $D$ to get $\theta_1, ..., \theta_{\lambda - 1}$
        \STATE Inject the actor: $\theta_{\lambda} \leftarrow \theta_a$

        \FOR{$\theta_i$ in $\theta_1, ..., \theta_{\lambda}$}
        \STATE Evaluate $\theta_i$ on $\mathcal{E}$ and store transitions in $R$
        \STATE Compute fitness $f_i$ of $\theta_i$
        \ENDFOR

        \STATE Update $D$ with the $\lambda$ individuals and their fitness values

        \FOR{$i=1$ to $N\_steps$}
        \STATE Sample $B$ transitions from $R$
        \STATE Update $\phi$ with a gradient step on the critic loss
        \STATE Update $\theta$ with a gradient step on the actor loss
        \ENDFOR
        \ENDFOR
    \end{algorithmic}
\end{algorithm}

\section{Genetic Drift}
\label{sec:drift}

We first characterize genetic drift in evolutionary methods, including a measurement on example policy search problems. We then discuss existing approaches to mitigate drift, before presenting \gdr{} in the next section.

\subsection{Detrimental genetic diversity}

In an ES, the population is encoded as a Gaussian distribution, which implies genomes are close to each other in parameter space at each generation. Injecting an actor that is far away in parameter space can therefore negatively impact the update step, which assumes that policies are normally distributed. We consider detrimental genetic diversity to be the inclusion of individuals in the ES update step which are outside of the population distribution.

The ES update uses a weighted sum of the policy parameters, using the previous distribution center as the starting point for the gradient estimate. With genomes $\theta_1 ... \theta_{N-1}$, ES center $\theta_c$, actor $\theta_a$, policy weights $w_i$, and actor weights $w_a$, the ES update is:

\begin{equation}
	\theta_c \leftarrow \theta_c + \sum_{i=1}^{N-1} w_i (\theta_i - \theta_c) + w_a (\theta_a - \theta_c)
\end{equation}

In this update, the ES distribution is moved based on a local gradient estimate using this weighted sum. The weight vector $w$ depends on the fitness rank of the policy; we use the same weight vector as in \cite{hansenCompletelyDerandomizedSelfAdaptation2001} and  \cite{chrabaszczBackBasicsBenchmarking2018}. We first note that that weights for the worst half of the population are $0$, hence the actor may have no impact on the update at all. When the actor performs well, however, its genes will be used to update the ES distribution, which it was not sampled from. Injecting the actor can therefore be detrimental as it can bring genes that are far away from the population and that do not properly inform the local gradient estimate.

The actor in these actor injection methods instead results from training in an RL algorithm like TD3. TD3 and SAC belong to the family of Value Iteration algorithms which optimize policies to maximize the Q-values of sampled states estimated through a critic network. The actor will be trained to follow the Q-value landscape, which is different from the true fitness landscape of the ES. At the beginning of training, the critic is not optimized and the Q-value landscape will not reflect the fitness landscape. It is therefore unlikely that actor training follows the same gradient direction as the ES update, which can lead the actor's genes to diverge from the ES population. Furthermore, the ES gradient is based in the parameter space directly, while the RL gradient uses backpropagation, leading to different updates even when gradient estimates are similar.

Some methods, such as ERL \cite{khadkaEvolutionGuidedPolicyGradient2018}, use actor injection in a genetic algorithm, and not an evolutionary strategy. In a genetic algorithm, a population of genetically diverse individuals are maintained throughout evolution, and there is no assumption that the population is close in parameter space. Injecting an actor with a genome very different from the population can even be beneficial, as it can bring new genes into the population for further mutation and crossover. However, ES have been demonstrated as highly capable algorithms for tackling the large and complex search space of neural network policy optimization \cite{chrabaszczBackBasicsBenchmarking2018}, so we focus in this work on mitigating the detrimental effects of genetic drift in ES.

\subsection{Measuring Genetic Drift in ES}

We define Genetic Distance as the Euclidean distance between the genomes of the actor $\theta_a$ and the ES center $\theta_c$ at a specific generation, $\lVert \theta_a - \theta_c \rVert$. %
Genetic distance is hence a way to measure the impact the actor could have on the ES update; even with a small weight, a large genetic distance will have a large impact on the update. When running an evolution strategy and training a TD3 from its transitions, we can measure the genetic distance at each generation. We show in Figure \ref{fig:drift} that without regularization, the genetic distance between the actor and the ES center increases with the number of generations. We call this phenomenon Genetic Drift.

\def\figwidth{0.3\linewidth}
\def\xleg{Evaluations}
\def\yleg{Distance}
\begin{table*}[h!]
	\centering
	\begin{tabular}{cccc}
		                                                                                       & \halfcheetah{} & \walker{} & \hopper{} \\
		\raisebox{5\normalbaselineskip}[1cm][0cm]{\rotatebox[origin=c]{90}{\vspace{1cm}\yleg}} &
		\includegraphics[width=\figwidth]{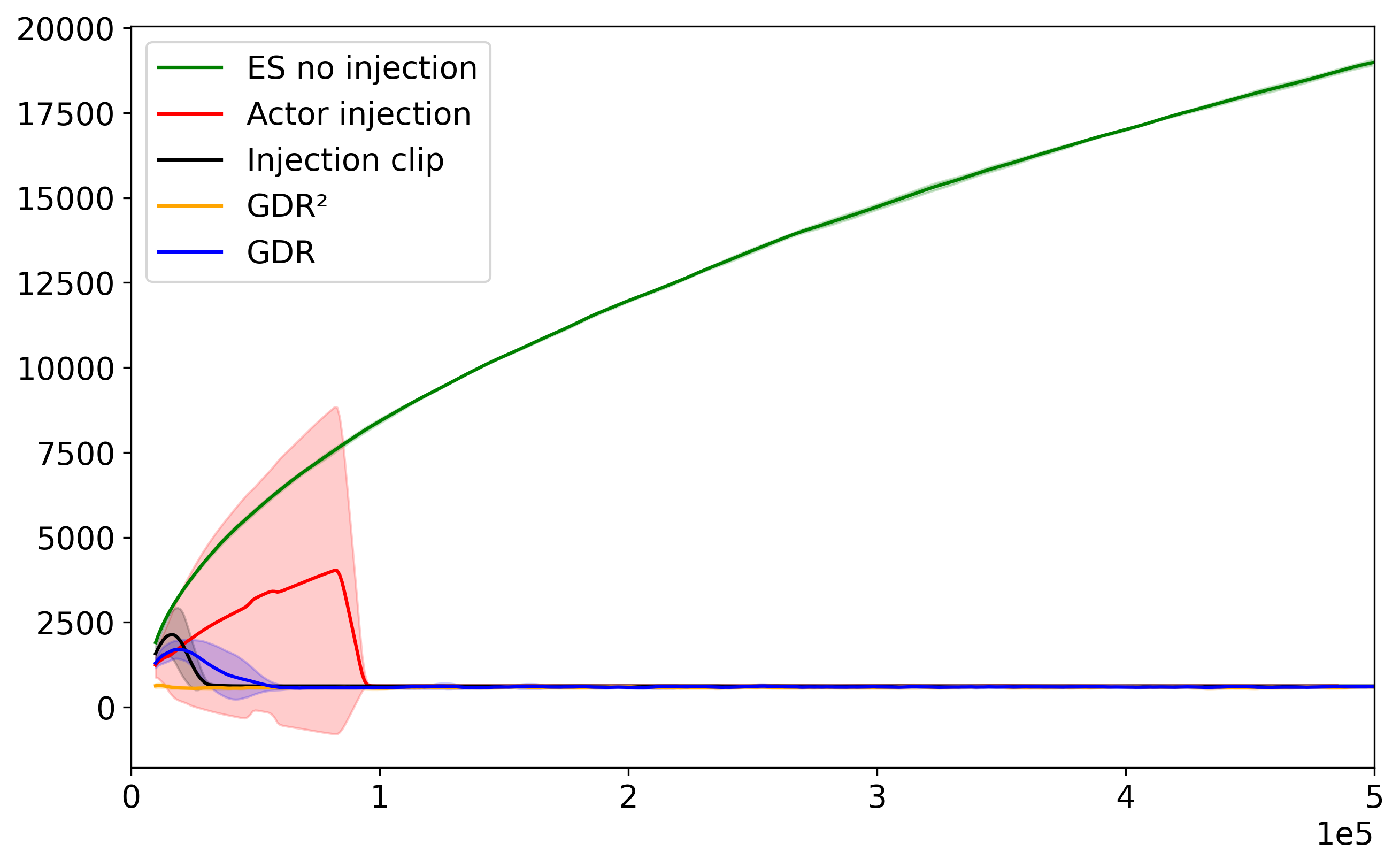}                     &
		\includegraphics[width=\figwidth]{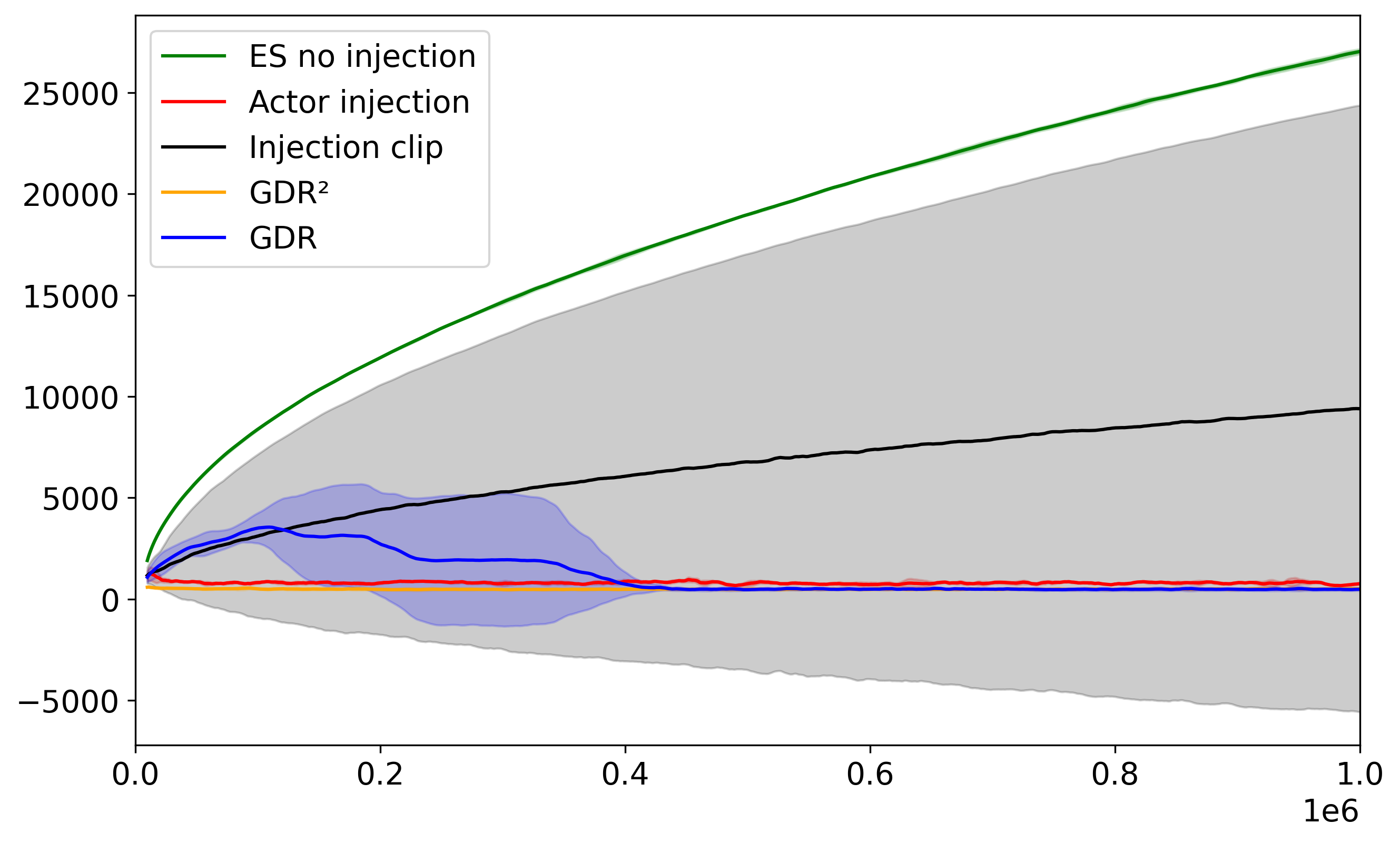}                        &
		\includegraphics[width=\figwidth]{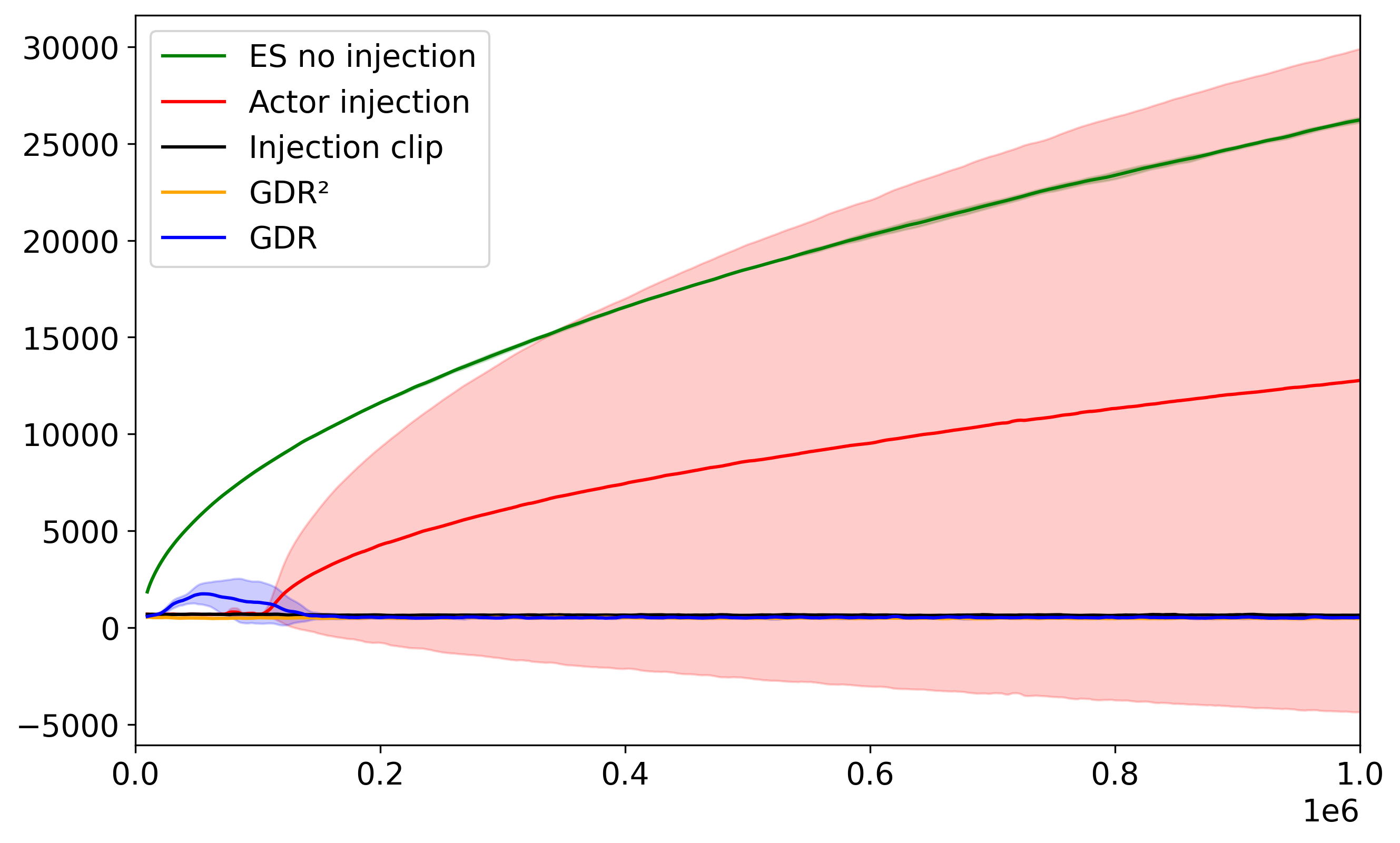}                                                                   \\
		                                                                                       & \xleg          & \xleg     & \xleg
	\end{tabular}
	\captionsetup{type=figure}
	\caption{Genetic drift: evolution of the genetic distance between genomes of the actor and the ES center.}
	\label{fig:drift}
\end{table*}

With actor injection, the genetic distance grows similarly when the actor is not good enough to be injected: the RL part has no impact on the ES and we get the same behavior as previously described. The ES and the actor start from the same point, but by the time the actor is injected, they have already moved away from each other: when the actor becomes good enough to be used in the update, its high genetic distance can hence make it strongly impact the ES update, pulling the ES center towards itself and reducing the genetic distance, as seen in \halfcheetah{}.

Actor injection acts as a regularization of the genetic drift by puling the ES towards the actor, but in the process often breaks the optimization process of the ES by making it change its search region. If the actor brings the ES distribution into a worst performing search area, it will keep performing well while the fitness of the rest of the population drops and hence pull the ES towards itself again. This explains the sharp drop in genetic distance seen on \halfcheetah{} with actor injection.

\subsection{Genetic drift mitigation}
The actor injection mechanism in ES has been studied directly in an approach aiming to improve ERL \cite{khadkaEvolutionGuidedPolicyGradient2018} by replacing the GA by an ES \cite{zhengRethinkingPopulationassistedOffpolicy2023}, and testing injection methods where the actor can be either ranked within the population, always injected, or always used as the best individual in the population. Results seem to depend on the environment considered, and the best method is not always the same. Performance of the actor is not reported, but it looks like always injecting the actor is good when the actor learns well, and injecting it normally (based on its rank in the population) is good when the actor is bad and hence ignored by the update.

One method to tackle issues in actor injection comes from the study of injecting external solutions into a CMA-ES population \cite{hansenInjectingExternalSolutions2011}. It clips the length of the vector used in the weighted sum so it does not have a disproportionate impact on the ES update, reducing the impact of the injected solution on the ES update. However in our case this method intervenes after the actor and the ES are already far apart from each other, and does not reduce genetic drift as seen on \walker{} in Figure \ref{fig:drift} plots, where this method is named "Injection clip". We use this method as a baseline in our experiments as it is still cited as a reference for actor injection by \cite{maheswaranathanGuidedEvolutionaryStrategies2019}.

As we consider the specific case of the injection of an actor optimized along the ES and both starting from the same initial genome, the two genomes are close enough in the early generations to not destabilize the ES with injection. We introduce the \longgdr{} method to make them stick together all along the optimization process, so they do not get to a point where injection might break the update.

\section{\longgdr{}}
\label{sec:method}

In order to keep the actor genetically close to the ES, we add a regularization in the actor training loss on its genetic distance to the center of the ES.

We consider a control problem in an environment to be defined by a Markov Decision Process \cite{bellmanMarkovianDecisionProcess1957} represented as a tuple $\mathcal{E} = \langle \mathcal{S}, \mathcal{A}, \mathcal{P}, \mathcal{R} \rangle$, where $\mathcal{S}$ is the set of states, $\mathcal{A}$ is the set of actions, $\mathcal{P}$ is the transition function from each state to the next one depending on the action, and $\mathcal{R}$ is the reward function. In general, we aim to find a policy $\pi: \mathcal{S} \rightarrow \mathcal{A}$ that maximizes the expected reward $\mathbb{E}_{\pi}[\mathcal{R}]$. In this case, we will do so through an ES and RL, injecting an actor from the RL method into the ES.

\subsection{Lagrangian formulation}
\label{sec:lagrangian}
We approach the problem as a constrained optimization problem where the actor is trained to minimize the standard actor loss while keeping the Euclidean distance to the center below a threshold $\delta$:

\begin{align*}
g(\pi_A) &= - \mathbb{E}_{s \sim \mathcal{B}} \left[ \hat{Q}(s, \pi_A(s)) \right]\\
h(\pi_A) &= \left\| \theta_A - \theta_{ES} \right\|_2 - \delta \leq 0
\end{align*}

with $\hat{Q}$ the state-action value function estimator, $\pi_A$ the actor policy with genome $\theta_A$, $\theta_{ES}$ the center of the ES, $\mathcal{B}$ the replay buffer and $s$ the states sampled from that buffer.

The Lagrangian of this problem is:

$$\mathcal{L}(\pi_A, \epsilon) = g(\pi_A) + \epsilon h(\pi_A)$$

To find the optimal values of $\pi_A$ and $\epsilon$, an iterative optimization process is used. At each iteration, the values of $\pi_A$ and $\epsilon$ are updated based on the current values. The update equations are as follows:

\begin{align*}
\pi_A^{k+1} &= argmin_{\pi_A} \mathcal{L}(\pi_A, \epsilon^k) \\
\epsilon^{k+1} &= argmax_{\epsilon} \mathcal{L}(\pi_A^{k+1}, \epsilon)
\end{align*}

These equations represent the update step for $\pi_A$ and $\epsilon$ in each iteration. The objective is to find the values of $\pi_A$ and $\epsilon$ that minimize the Lagrangian function while satisfying the constraint $\epsilon \geq 0$.

\subsection{Lagrangian relaxation}

To simplify the method we relax the constraint by simply adding the distance to the loss with a fixed scaling parameter $\epsilon$. 
$\hat{g}_{GDR}(\pi_A)$ is the regularized loss minimized during the actor training steps to still maximize expected Q-values while staying close to the ES center. We call this method \longgdr{} (\gdr{}) since it directly aims to reduce the genetic drift between the ES center and the actor.

$$\hat{g}_{GDR}(\pi_A) = - \mathbb{E}_{s \sim \mathcal{B}} \left[ \hat{Q}(s, \pi_A(s)) \right] + \epsilon \left\| \theta_A - \theta_{ES} \right\|_2  $$

\subsection{Squared distance}

In the ES update the Euclidean distance between the center of the ES and an agent can be seen as the scaling factor of its unit vector in the update direction, hence the L2 distance has a direct relationship to the ES update which motivates using it as regularization. Since all norms are equivalent in finite dimensions, any other norm or monotonous transformation of a norm could replace it. We can generalize the regularized objective from \gdr{} to any other convex regularization term, such as the square of the L2 norm. We name that method \longsgdr{} (\sgdr{}):

$$\hat{g}_{SGDR}(\pi_A) = - \mathbb{E}_{s \sim \mathcal{B}} \left[ \hat{Q}(s, \pi_A(s)) \right] + \epsilon \left\| \theta_A - \theta_{ES} \right\|_2^2  $$

Other regularization terms are not tested here since they would need to be related to the ES update mechanism, which remains based on summing vectors in the genome space. The added computational cost of computing the L2 norm in the loss is negligible compared to the rest of the computation, making \gdr{} virtually free to add to any method with actor injection.

\section{Experiments and Results}
\label{sec:results}

In this article we isolate the actor injection mechanism to study its impact on the behavior of the ES center and on the RL actor: GDR on itself is not a direct competitor to full methods that also use gradient information from the critic. While recent methods such as ERL-Re² \cite{hao2022erl} or Self-Guided ES \cite{liuSelfGuidedEvolutionStrategies2020} can yield more state-of-the art results in evolutionary RL, their many other components or lack of actor injection at all make them out of the scope of this work. This work focuses on evaluating the impact of genetic drift and \gdr{} in a simpler setting, comparing it to standard injection methods, and we leave for future work the study of how \gdr{} can be used with ERL-Re² or CEM-ACER compared to standard injection. 

\subsection{Control tasks}

We benchmark our methods on the Brax control tasks \cite{brax2021github} with continuous actions. Each rollout is 1000 steps long, with the same initial state for all evaluations. Because the rollouts are run on GPU, there is still some stochasticity even with (theoretically) deterministic policies and environment updates due to GPU operation stochasticity which propagate along the 1000 steps.

Each agent is evaluated once per generation, then every 10 generations the ES distribution center is evaluated 100 times and the average is logged as the ES fitness. Each evaluation is a full rollout of the policy with no exploration noise for the RL actor.

\subsection{Algorithms}

To study the impact of actor injection and \longgdr{} (\gdr{}), we compare the following algorithms:
\begin{itemize}
    \item Canonical ES: the canonical ES algorithm \cite{chrabaszczBackBasicsBenchmarking2018} , with no actor injection. To measure actor performance, we train a TD3 actor from the samples collected during evolution to observe the impact of actor injection on the actor performance, but this has no impact on the ES. 
    \item Actor Injection: the canonical ES algorithm with actor injection. The actor is trained with TD3 on samples from the evaluation of ES agents (including the actor).
    \item Injection clip: Canonical ES with actor injection. We add the clipping mechanism introduced in \cite{hansenInjectingExternalSolutions2011} to limit the impact of the actor on the update when its genome is too far from the ES population.
    \item Parallel TD3: TD3 with full rollouts of the actor in parallel, and gradient steps at the end of the rollouts phase, described below.
    \item \gdr{}: Canonical ES with actor injection and \longgdr{}. The actor is trained with TD3 on samples from the evaluation of ES agents (including the actor).
    \item \sgdr{}: Canonical ES with actor injection and \longsgdr{}
\end{itemize}

\subsubsection{Parallel TD3}

We implement a parallel version of TD3 where the actor does multiple rollouts in parallel without training steps, then the critic and actor are trained using the transitions stored in the replay buffer from the rollouts. Although quite different from the TD3 method \cite{fujimotoAddressingFunctionApproximation2018}, this gives a baseline with similar architecture to the ES + RL methods in order to compare fairly. The full pseudocode is described in Appendix \ref{app:td3_code},

With the same number of rollouts as an ES, this parallel TD3 effectively does the same number of gradient steps as the ES + RL framework but without the ES learning strategy. This allows for the independent comparison of the evolutionary and reinforcement learning update steps.

Since there is no exploration noise in the genome space from an ES, we add exploration noise to the actor actions during the rollouts phase, as is common in RL. This noise is sampled from a normal distribution with mean 0 then added to the action.

\subsubsection{Hyperparameters}

Base parameters used in our experiments are described in Table \ref{table:hyperparameters} in Appendix \ref{app:code}. Hyperparameters were picked based on the default hyperparameters for mixing evolution and RL in QDax, then tested with a grid search on a log scale for hyperparameters with at least 3 values (typically testing 0.1, 1, 10 for the ES $\sigma$). We then use the same parameters for all combinations of the two methods. Final parameters were picked based on final performance and stability of the fitness of the ES center.

We run each experiment to $10^6$ evaluations on all environments, except on \halfcheetah{} which we stop at $5.10^5$ evaluations as all methods tend to plateau early. Experiments on \humanoid{} are stopped after 24h, which due to more expensive evaluations led to plots being stopped at $6.10^5$ evaluations.

Policy and critic networks architectures have 2 hidden layers of 128 neurons, in addition to input and output layers which depend on the environment. Activation functions are ReLU between all layers, with an additional tanh activation on the output of the actor. Although this architecture yields lower scores on pure ES optimization than the 4 hidden layers of 32 units with tanh used in evosax \cite{langeEvosaxJAXbasedEvolution2022}, the tanh activation in a deep narrow network can lead to vanishing gradients and tends to make it harder for the RL actor to learn. Since we use the same architecture for both ES agents and the RL actor to inject its genome, we use the same architecture for all agents and consider the architecture with 2 layers of 128 and ReLU as part of the optimization problem.

For the \gdr{} and \sgdr{} $\epsilon$ value, we test 3 values at 0.1, 0.01 and 0.001 over 5 runs and pick the best performing for each environment. Values used in results here are shown in Table \ref{table:epsilon_hp} in Appendix \ref{app:code}, and we show learning curves for all values in Figure \ref{table:epsilon}.

\def\figwidth{0.3\linewidth}
\def\xleg{Evaluations}
\def\yleg{Fitness}
\begin{table*}[h!]
    \centering
    \begin{tabular}{cccc}
                                                                                                       & \halfcheetah{} & \walker{} & \hopper{} \\
        \raisebox{5\normalbaselineskip}[1cm][0cm]{\rotatebox[origin=c]{90}{\vspace{1cm}ES fitness}}    &
        \includegraphics[width=\figwidth]{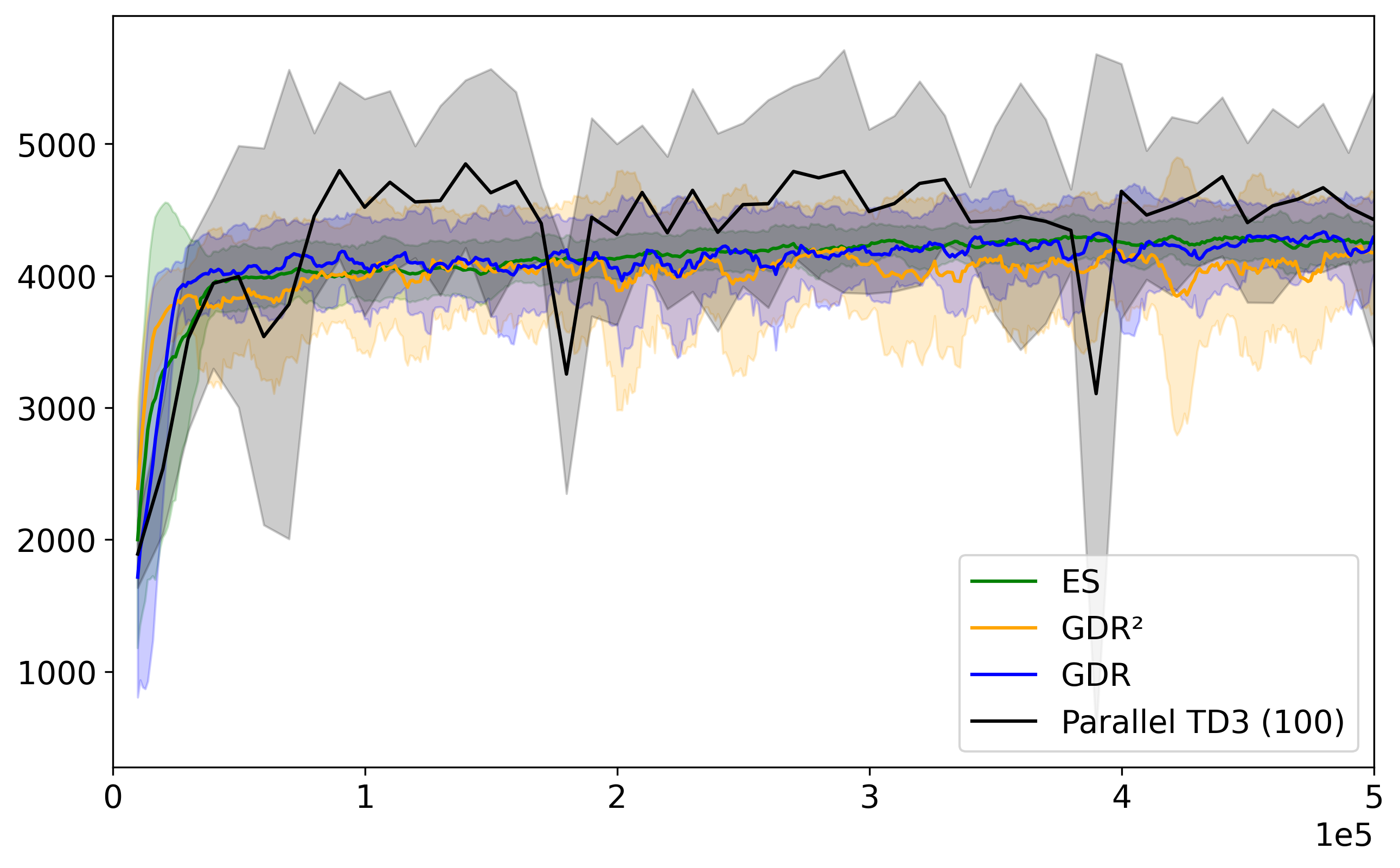}                         &
        \includegraphics[width=\figwidth]{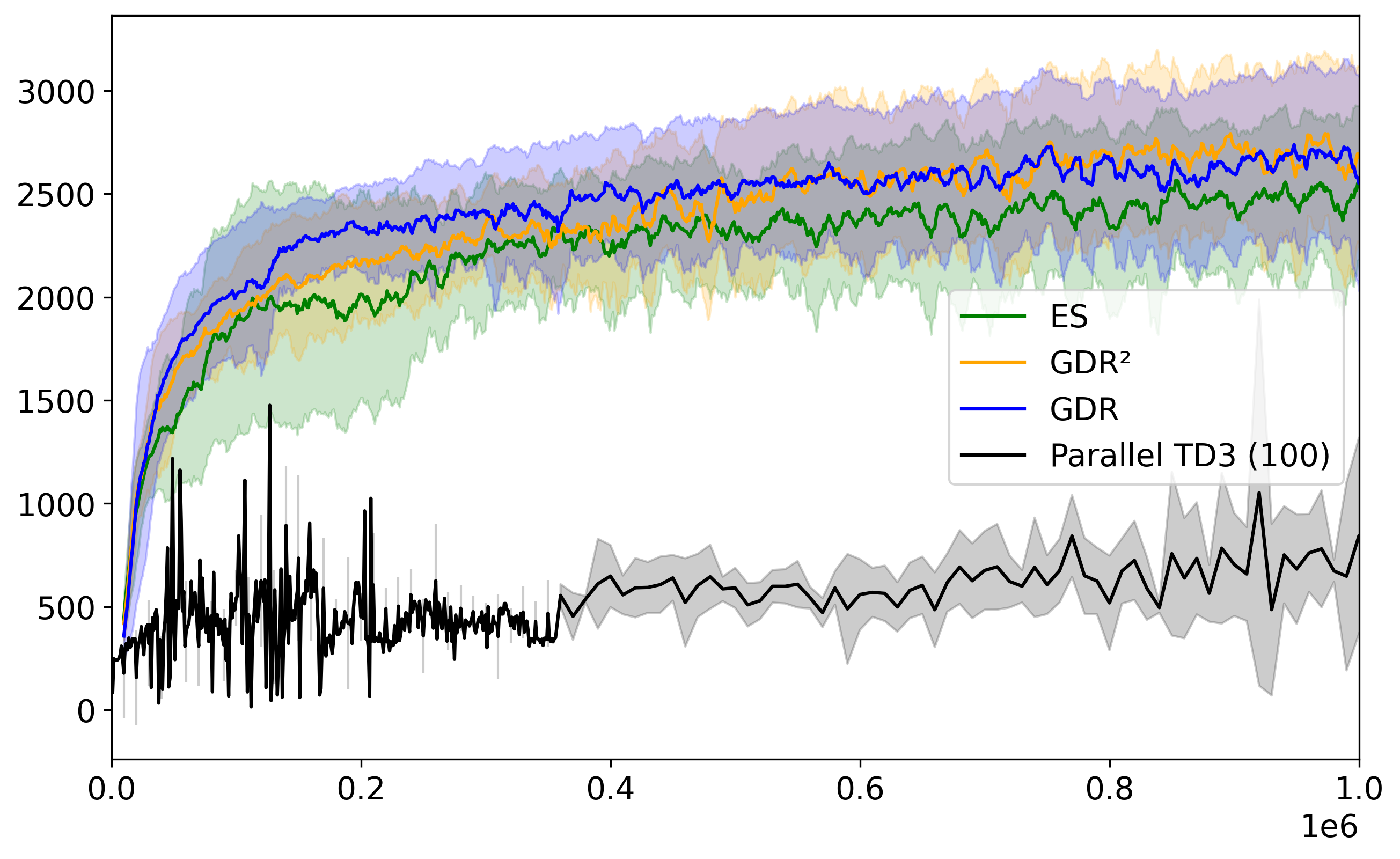}                            &
        \includegraphics[width=\figwidth]{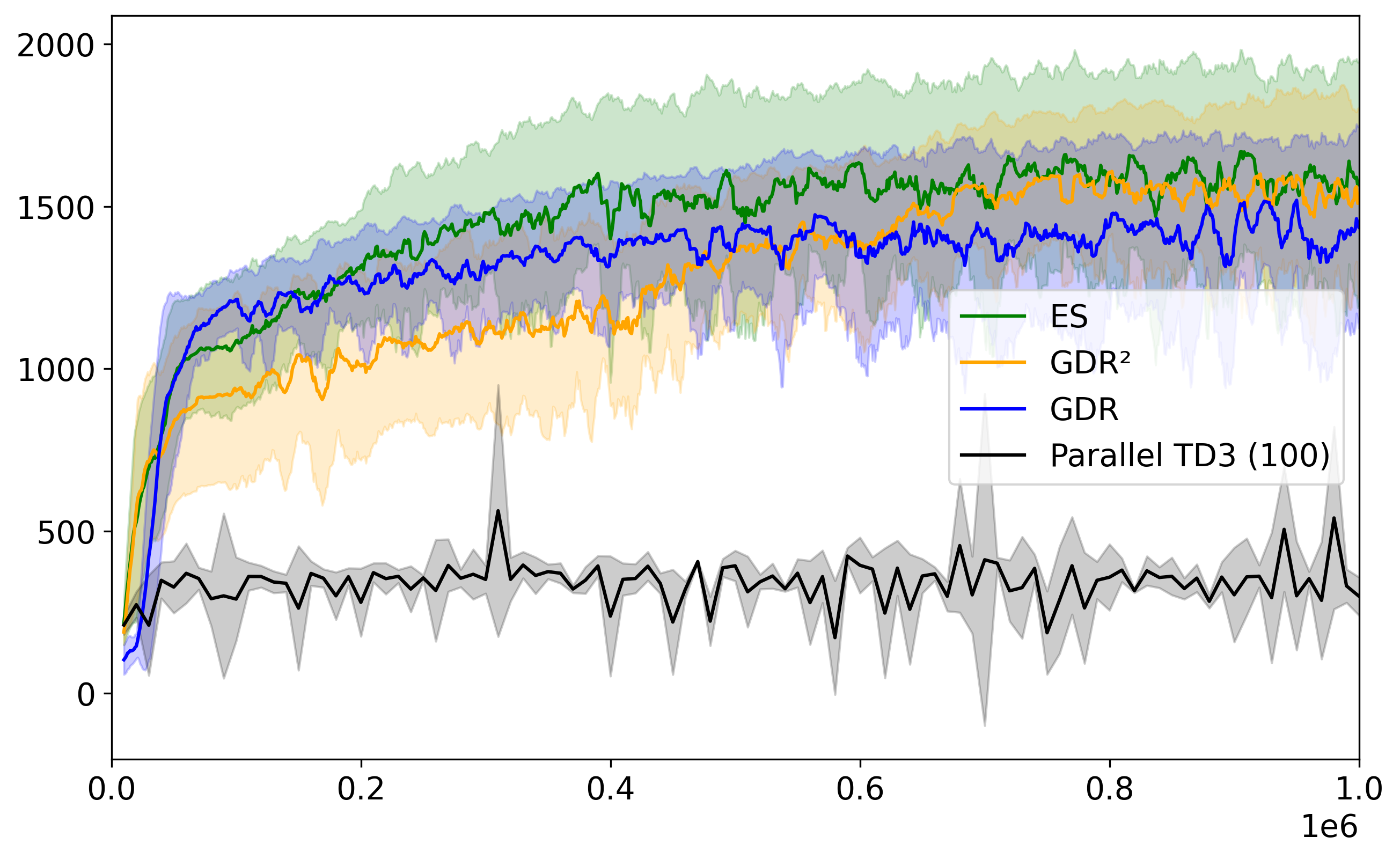}                                                                       \\

        \raisebox{5\normalbaselineskip}[1cm][0cm]{\rotatebox[origin=c]{90}{\vspace{1cm}ES fitness}}    &
        \includegraphics[width=\figwidth]{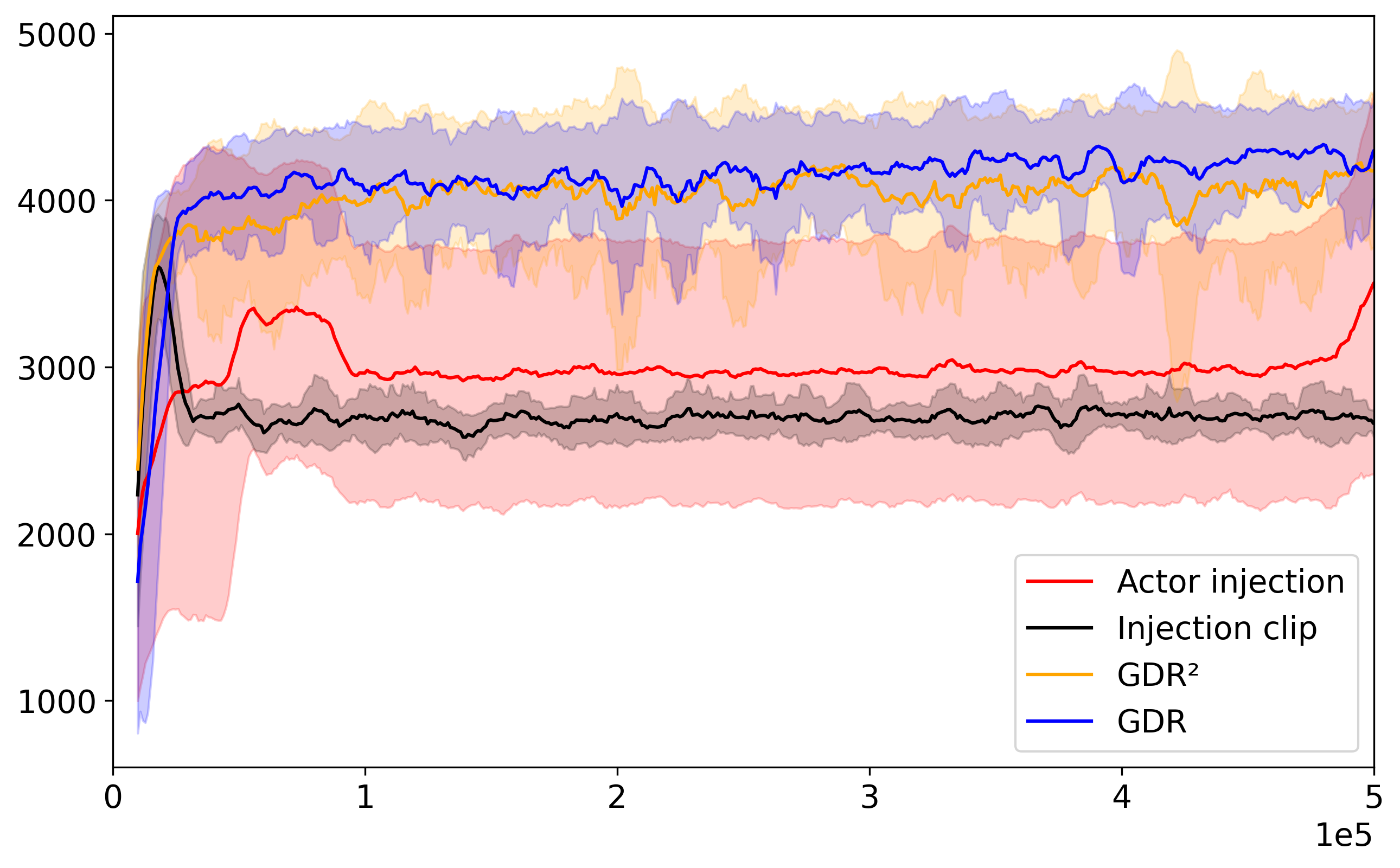}                         &
        \includegraphics[width=\figwidth]{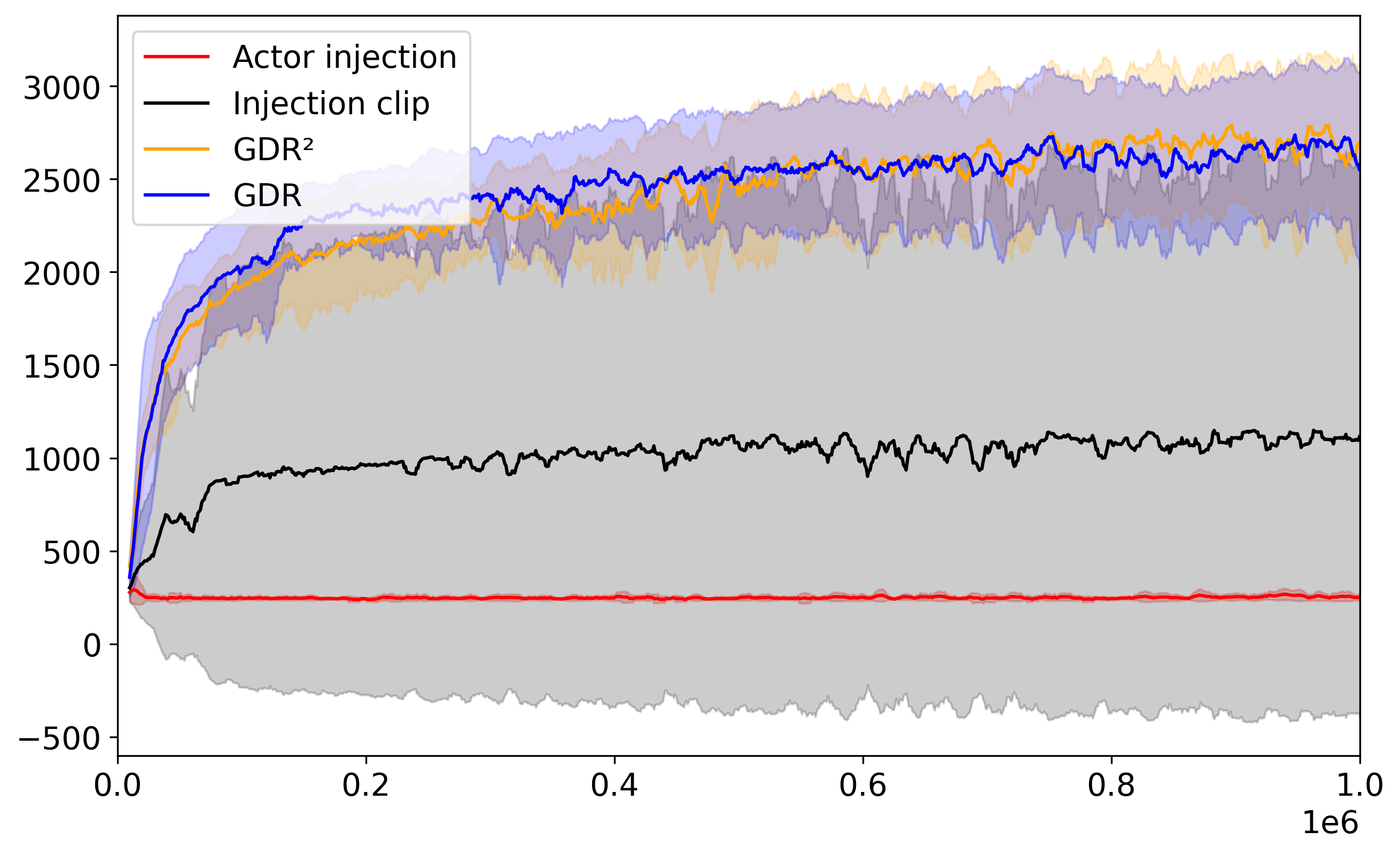}                            &
        \includegraphics[width=\figwidth]{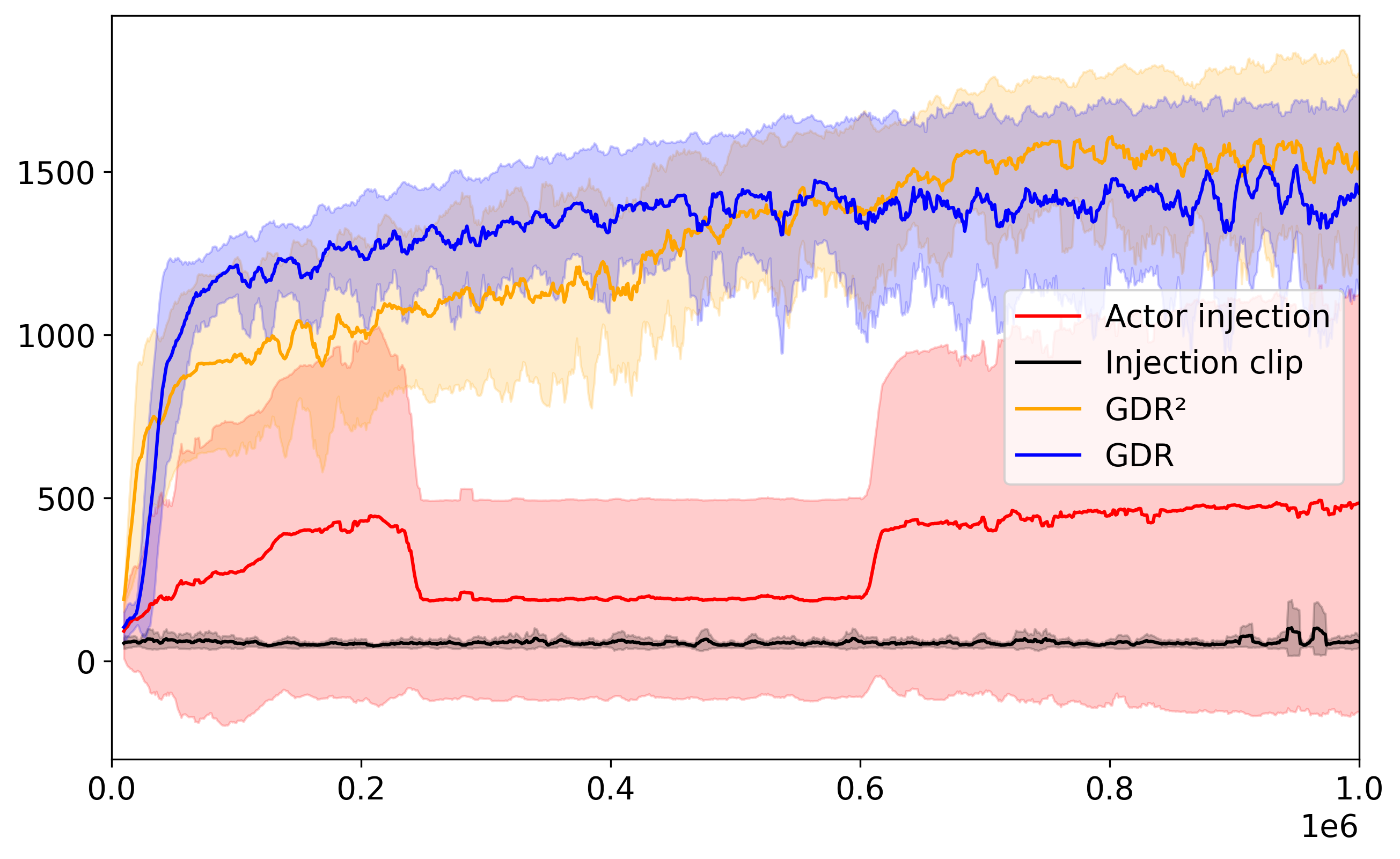}                                                                       \\

        \raisebox{5\normalbaselineskip}[1cm][0cm]{\rotatebox[origin=c]{90}{\vspace{1cm}Actor fitness}} &
        \includegraphics[width=\figwidth]{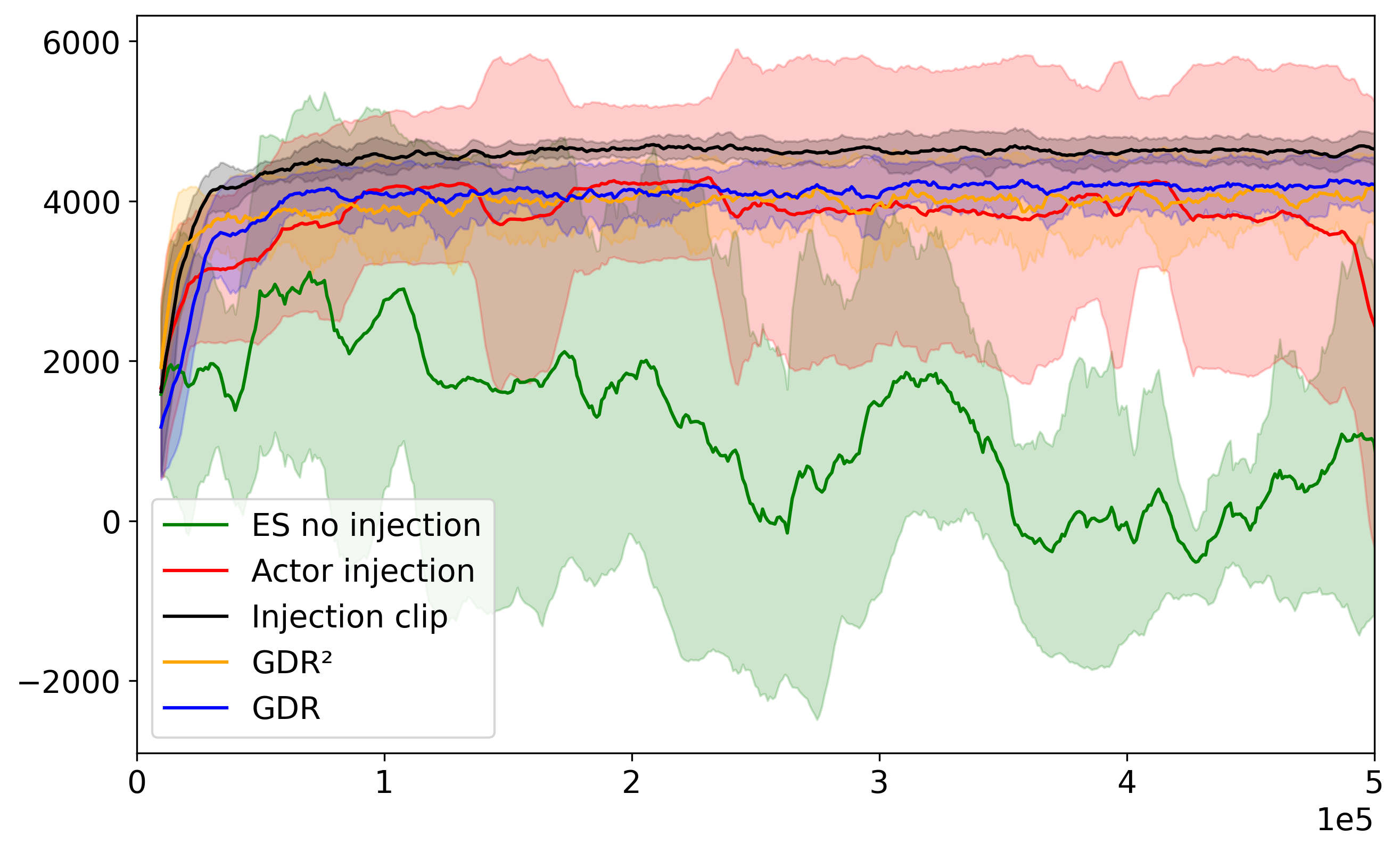}                             &
        \includegraphics[width=\figwidth]{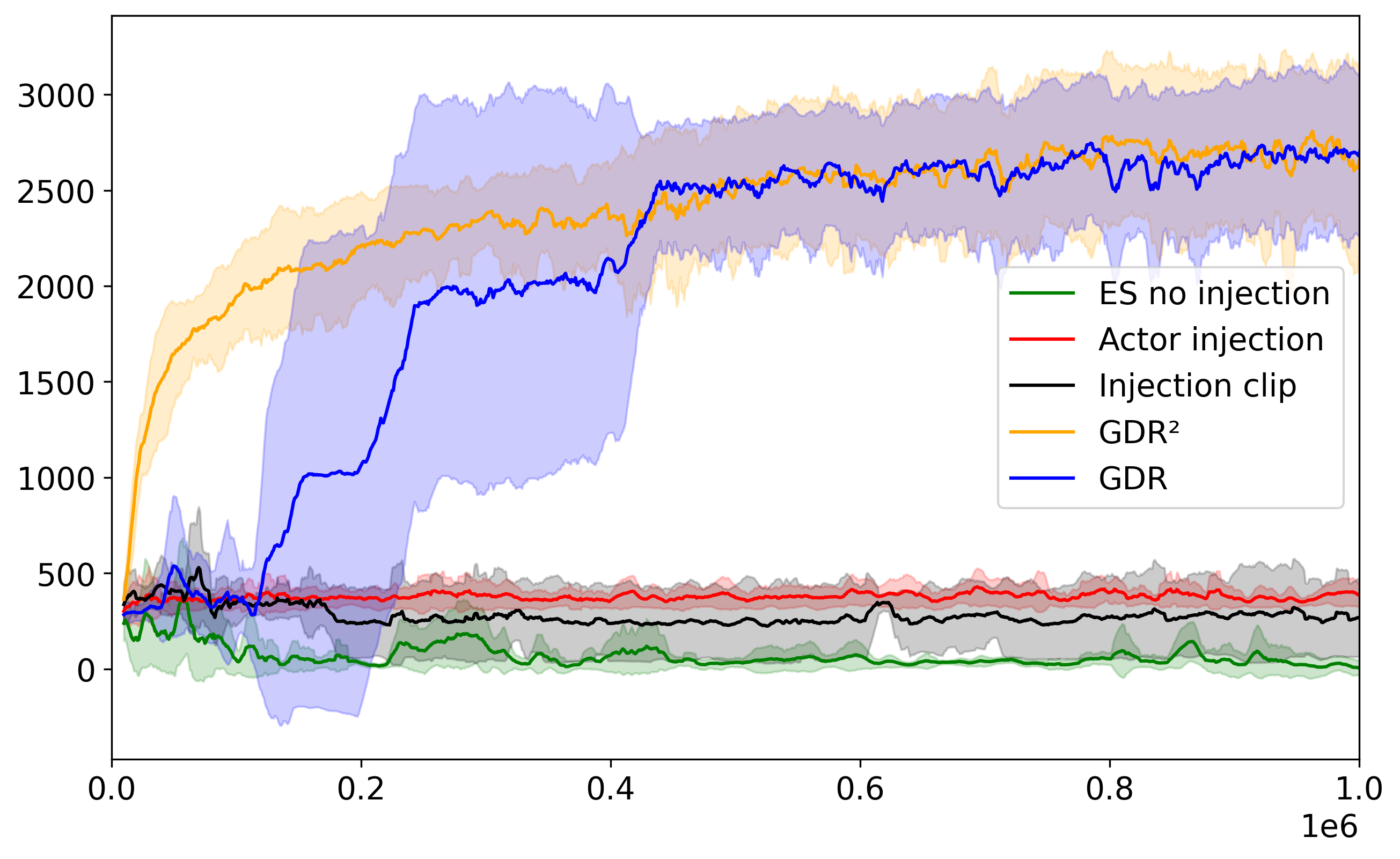}                                &
        \includegraphics[width=\figwidth]{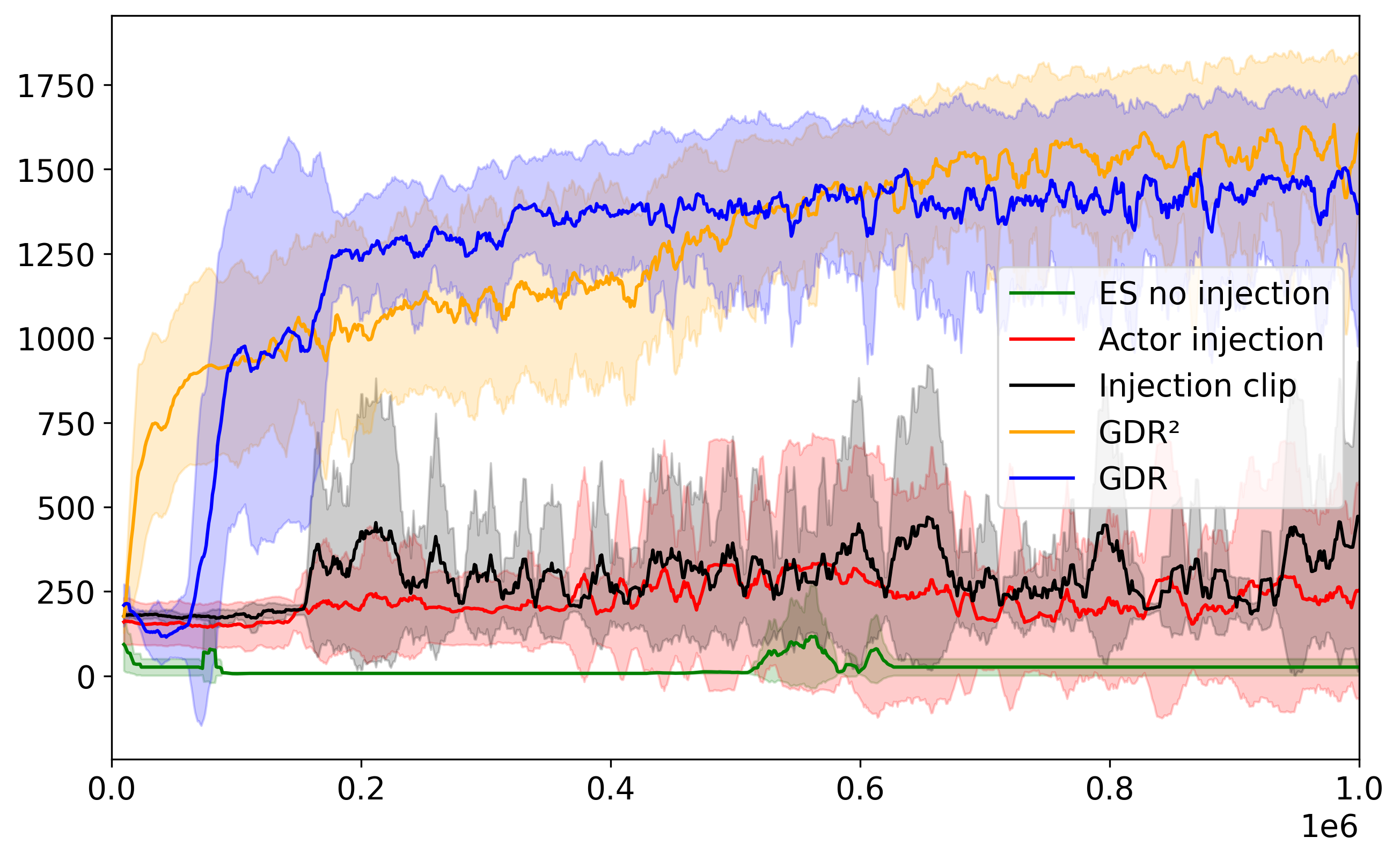}                                                                           \\
                                                                                                       & \xleg          & \xleg     & \xleg
    \end{tabular}
    \captionsetup{type=figure}
    \caption{Fitness of the ES center and the RL actor during training on 3 tasks. First and second rows compare \gdr{} and \sgdr{} to no-injection baselines and injection baselines respectively for readability. Third row shows the fitness of the actor during training. Colored areas show plus and minus one standard deviation.}
    \label{plot:results}
\end{table*}

    \begin{table*}[h!]
    \centering
    \begin{tabular}{l|cccccc}
 & \halfcheetah{} & \walker{} & \hopper{} & \swimmer{} & \humanoid{} & \ant{}\\
\hline
ES & 4254 (129) & \underline{2554} (375) & \textbf{1570} (376) & 108 (0) & 661 (178) & -9 (6)\\
Parallel TD3 (100) & \textbf{4427} (961) & 844 (475) & 299 (59) & -1 (2) & 85 & \textbf{2583} (1465)\\
Actor injection & 3501 (1144) & 249 (17) & 484 (641) & \textbf{109} (0) & 558 & -780 (831)\\
Injection clip & 2661 (78) & 1117 (1506) & 58 (18) & \underline{109} (0) & 694 (104) & -1058 (851)\\
GDR & \underline{4296} (286) & 2549 (519) & 1434 (310) & 109 (0) & \textbf{741} (83) & \underline{22} (75)\\
GDR² & 4175 (457) & \textbf{2679} (440) & \underline{1511} (291) & 109 (0) & \underline{708} (160) & -4 (0)\\
\end{tabular}
\caption{Average return of the final policy found by each algorithm on each environment. Format: mean (std), best score in bold and second best underlined.}
\label{tab:results}
\end{table*}

\subsection{Results}

\subsubsection{Genetic drift reduction}

In Figure \ref{fig:drift}, we see the genetic drift is reduced with \gdr{} and \sgdr{}, as anticipated. On \halfcheetah{} and \walker{}, the distance first grows before decreasing without the steep drop observed with standard actor injection.

\subsubsection{ES policy search}
We run the ES with and without actor injection, and compare the performance of the policy at the ES center. We also compare the performance of the ES with and without \gdr{}. Results are shown in Figure \ref{plot:results} and final fitness values are reported in Table \ref{tab:results}. The negative effect of actor injection can be noted in \halfcheetah{}, \walker{}, \hopper{}, and \ant{}, where the ES performs better without actor injection than with it. The impact of injection clipping is inconsistent but yields results close to standard actor injection. \gdr{} achieves similar performance to the better of the ES or TD3 on all environments except for \ant{}, where only TD3 learns an effective policy, and improves over standard actor injection on all environments.

\subsubsection{Actor training}

While \gdr{} directly impacts the loss of the actor and hence its training, the RL algorithm is still able to find good actor policies with genomes close to the ES distribution. Furthermore, on environments where the parallel TD3 method fails and ES + RL with no injection does not find a good actor, for example \walker{} and \hopper{}, \gdr{} shows a strong learning curve for the actor, bringing it along the ES optimization path to better search areas. The injection of this better actor can lead to improved ES performance as seen on \walker{}.

\subsection{Parameter sensitivity}
\def\figwidth{0.3\linewidth}
\def\xleg{Evaluations}
\def\yleg{Fitness}
\begin{table*}[h!]
    \centering
    \begin{tabular}{cccc}
                                                                                                       & \halfcheetah{} & \walker{} & \hopper{} \\
        \raisebox{5\normalbaselineskip}[1cm][0cm]{\rotatebox[origin=c]{90}{\vspace{1cm}ES fitness}}    &
        \includegraphics[width=\figwidth]{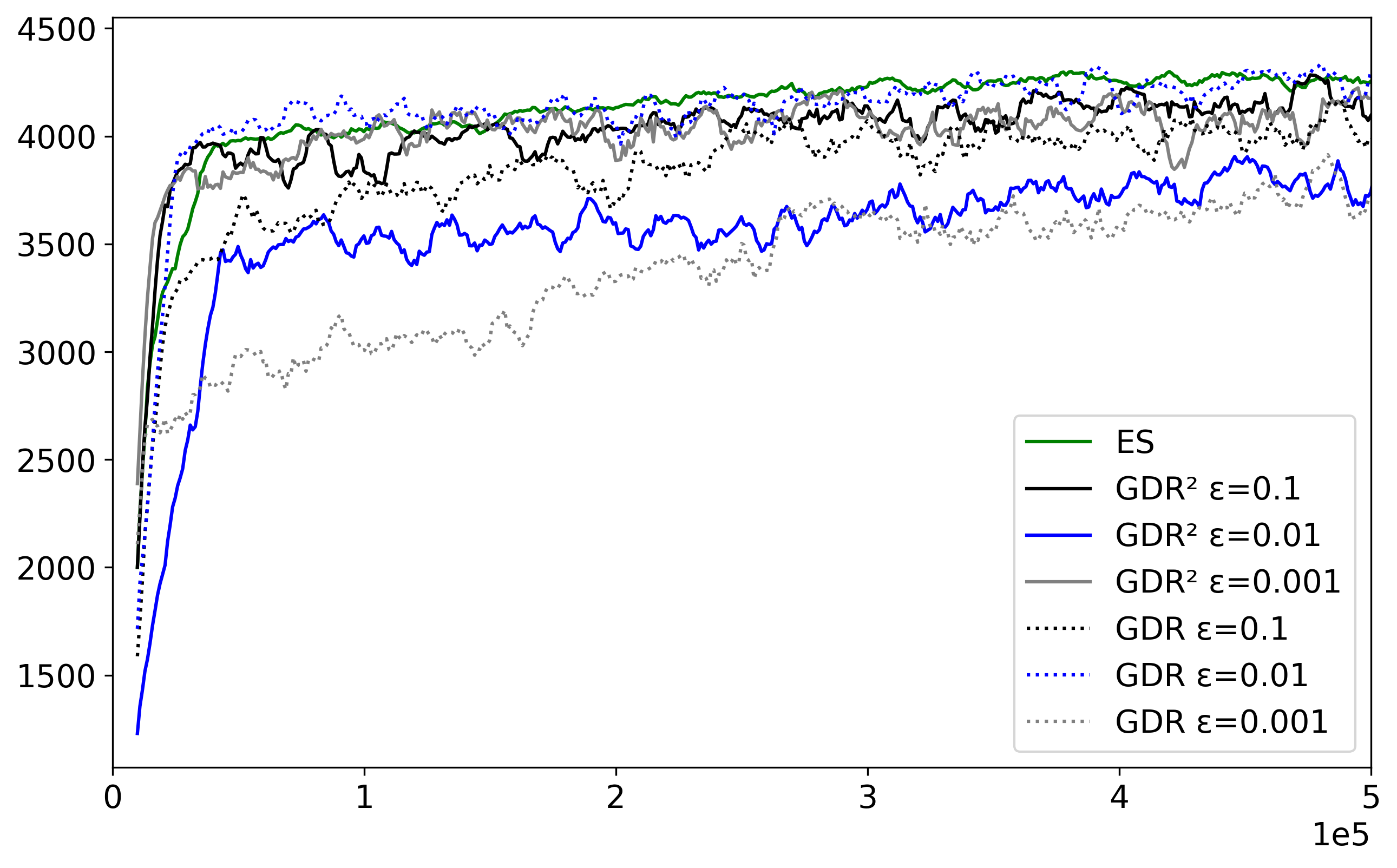}                          &
        \includegraphics[width=\figwidth]{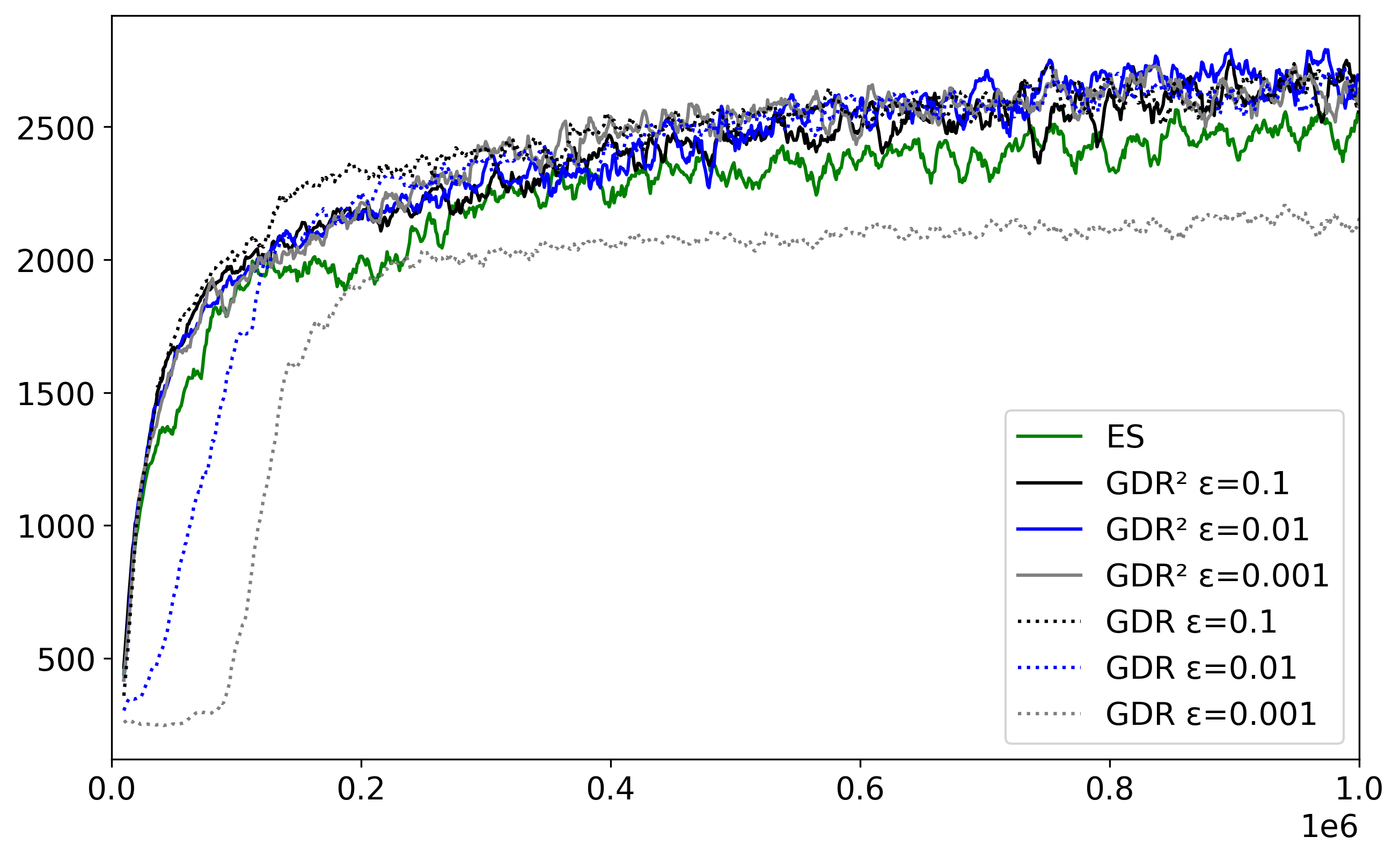}                             &
        \includegraphics[width=\figwidth]{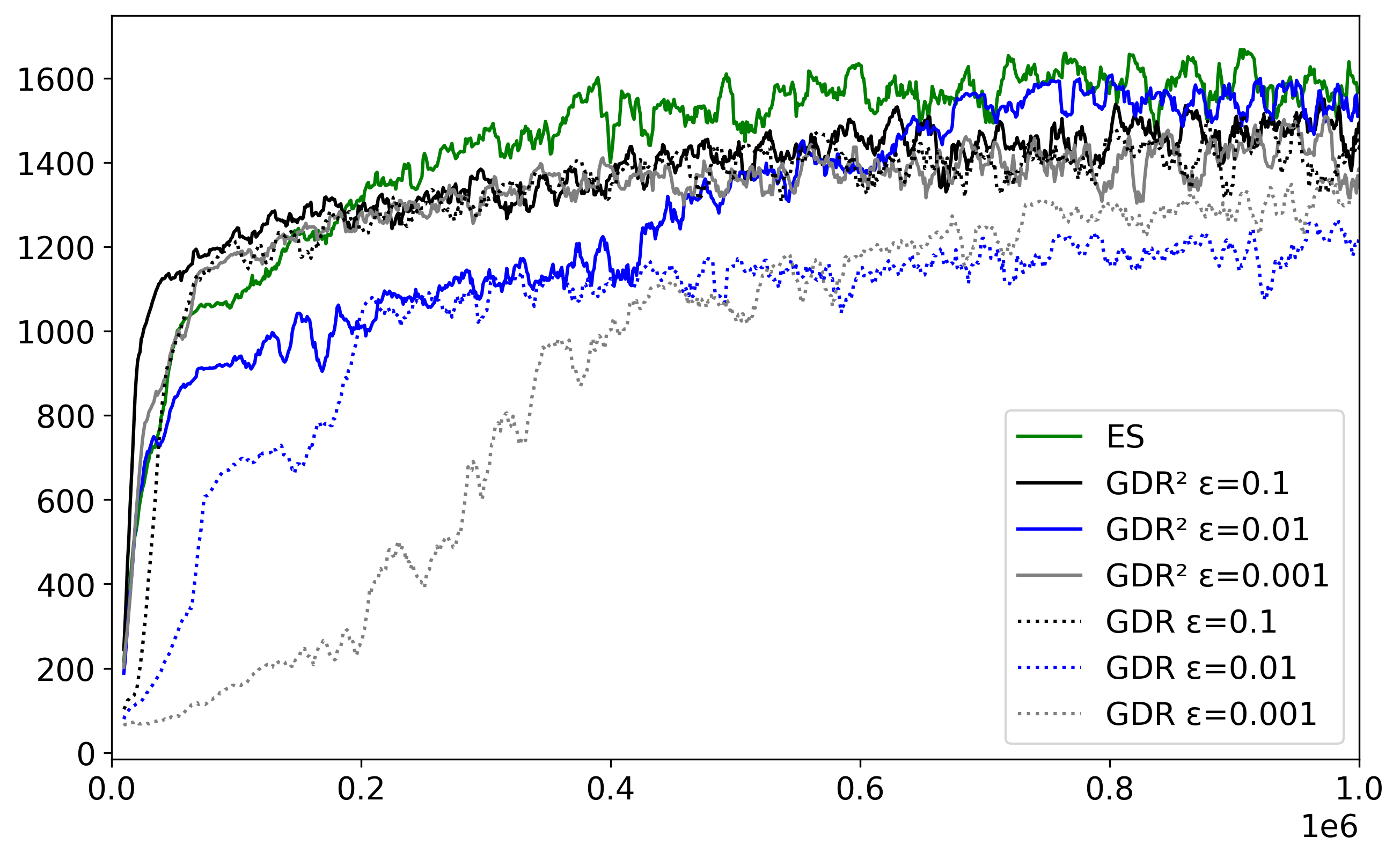}                                                                        \\

        \raisebox{5\normalbaselineskip}[1cm][0cm]{\rotatebox[origin=c]{90}{\vspace{1cm}Actor fitness}} &
        \includegraphics[width=\figwidth]{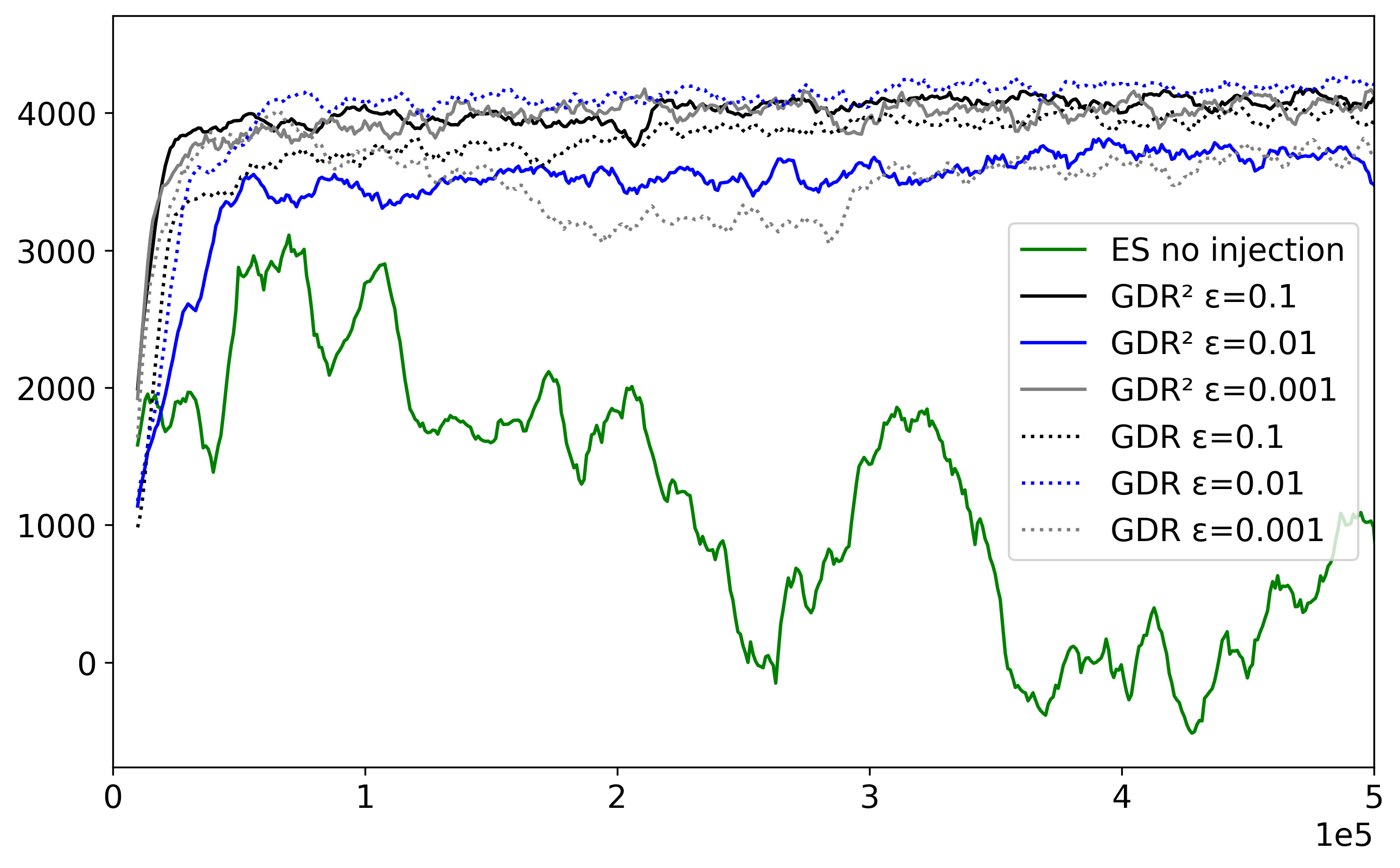}                         &
        \includegraphics[width=\figwidth]{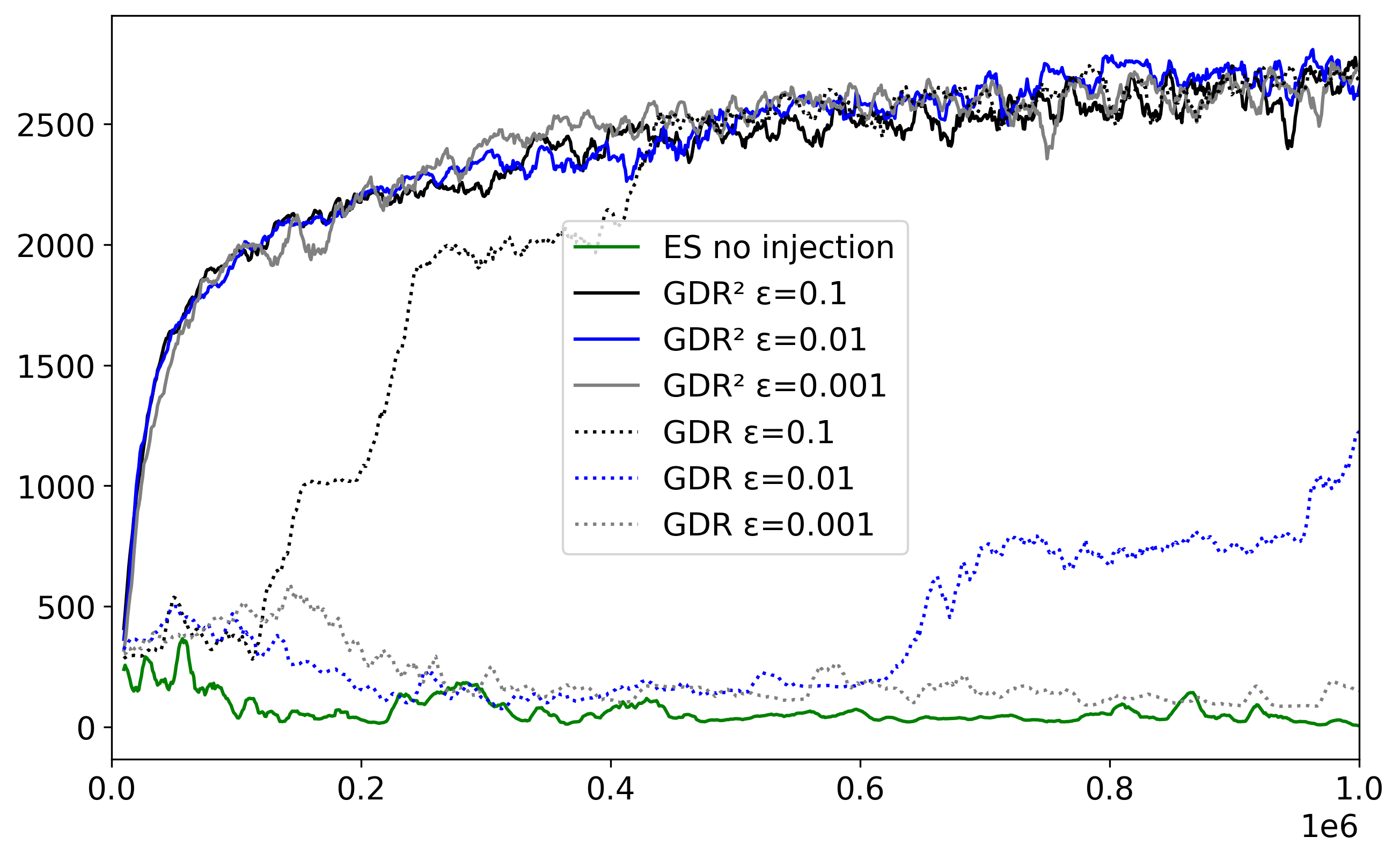}                            &
        \includegraphics[width=\figwidth]{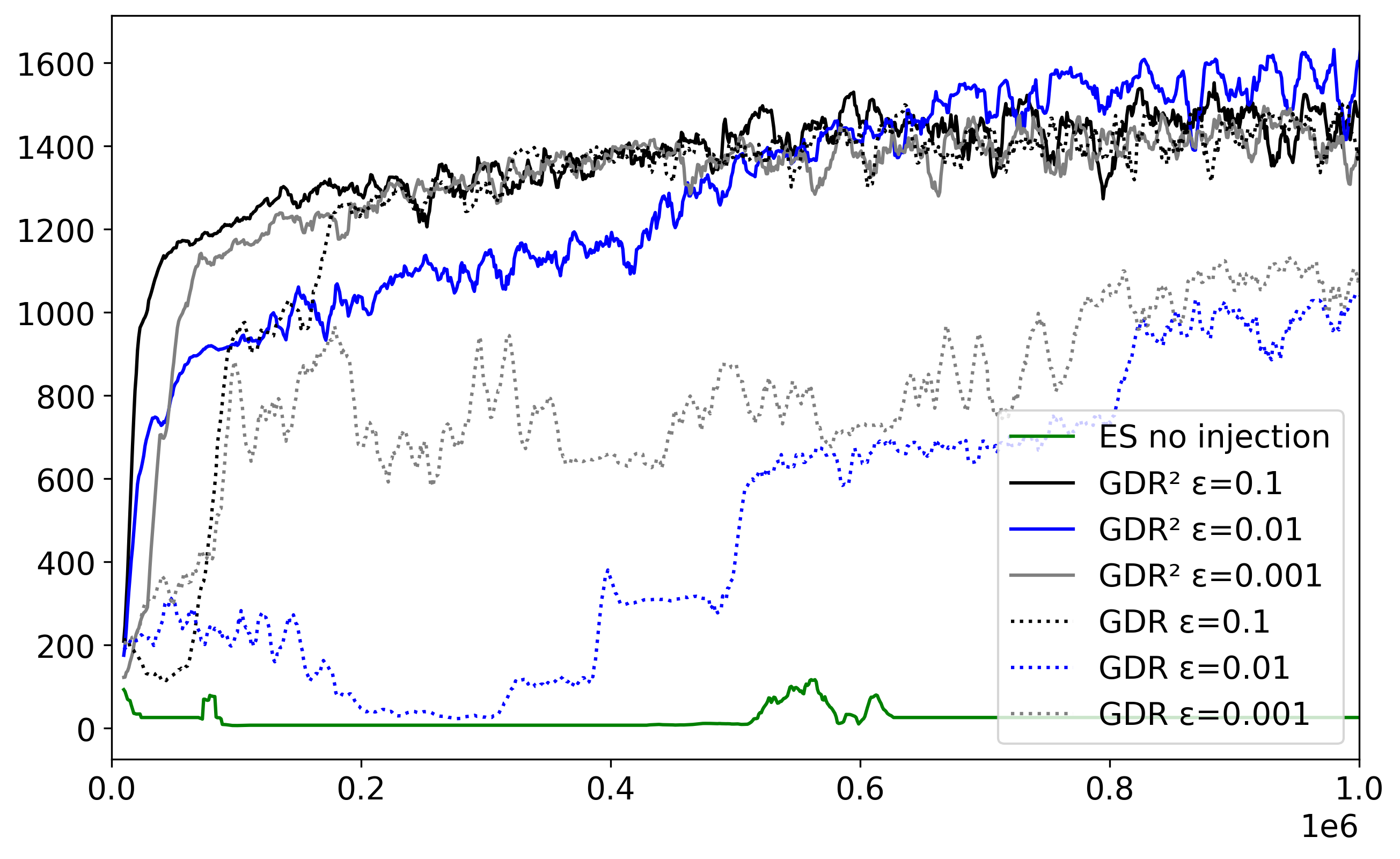}                                                                       \\
                                                                                                       & \xleg          & \xleg     & \xleg
    \end{tabular}
    \captionsetup{type=figure}
    \caption{Fitness of the ES center during training for different values of $\epsilon$ in \gdr{} and \sgdr{}.}
    \label{table:epsilon}
\end{table*}

We study the sensitivity of \gdr{} to different values of the regularization factor $\epsilon$ in Figure \ref{table:epsilon} plots and final scores are available in Appendix \ref{app:results}.

We see that although the tested values of $\epsilon$ have an impact on the final fitness of the ES center, none of them break the learning curve and both \gdr{} and \sgdr{} are quite resilient to changes of $\epsilon$ on \walker{}. With both regularization methods, the actor learning curve is improved on all 3 environments shown for most of the $\epsilon$ values over standard injection or simply RL trained on ES transitions with no injection. This shows that \gdr{} can still be effective even with $\epsilon$ not tuned, reducing the need for extensive hyperparameter search to reach decent results.

\section{Future works and conclusions}
\label{sec:conclusion}

\subsection{Future work on \gdr}

Although a direct regularization in the actor loss already alleviates the issue of genetic drift, there are still many ways to improve the method and its application to other problems.

First, the Lagrangian relaxation could be replaced by the computation of the trust region, as defined in section \ref{sec:lagrangian}, for a more complex but more accurate constrained optimization method.
Second, the method could be combined with other methods such as CEM-ACER \cite{tangGuidingEvolutionaryStrategies2021} to use the TD3 critic gradient to guide the ES towards better solutions and see if improved actor injection can help the ES converge faster.
Finally, the method could be applied to more complex problems such as discrete action or image-based control like MinAtar \cite{youngMinAtarAtariInspiredTestbed2019}.%

\subsection{Aligning landscapes}

While TD3 and SAC have been used in combination with ES, they are both off-policy methods which allow for the use of transitions from different policies in the same replay buffer, which enables the use of the ES population rollouts in the RL training loop. In order to better match the optimization landscapes and reduce drift, it could be interesting to use a policy gradient method such as Proximal Policy Optimization (PPO, \cite{schulmanProximalPolicyOptimization2017}) which optimizes policies towards the policy gradient landscape. This would make the actor follow the same landscape as the ES, and could reduce the drift between the two. PPO is an on-policy method, which means that the samples from the ES population rollouts would not be usable in the RL training loop, but recent work has introduced an off-policy version of PPO \cite{Meng_Zheng_Pan_Yin_2023}. We suggest \gdr{} could be used to help the RL learn from the ES, and actor injection for the ES to learn from the RL actor.

\subsection{Conclusions}

This study on actor injection in evolution strategies is a step in the direction of better understanding the interactions between evolution and reinforcement learning.
While value iteration and evolution can be complementary, they optimize on different landscapes leading to the actor and the ES tending to drift away from each other hence making actor injection potentially detrimental to the ES. \longgdr{} introduces a simple regularization with practically no additional cost to reduce this effect from the first generation, exploiting symmetries in the search space to allow the RL method to still find good actors with genomes close to the ES. This constraint on the actor can also help it find good areas of the search space, hence helping the RL part of the global method which in turn can improve the convergence of the ES.

\nopagebreak
\bibliography{references}
\nopagebreak
\appendix
\section{Appendix}
\subsection{Reproducibility}
\label{app:code}

Experiments were implemented using the QDax framework \cite{lim2022accelerated} and the \gdr{} codebase is made available\footnote{https://anonymous.4open.science/r/GDR-E784}. Each experiment was run 5 times with different random seeds picked as the first positive integers and all runs were used to compute mean and standard deviations. 
Hyperparameters are detailed in Tables \ref{table:hyperparameters} and \ref{table:epsilon_hp}. 

\begin{table}[h!]
    \centering
    \begin{subtable}{0.45\linewidth}
    \centering
    \begin{tabular}{c|c}
        Parameter                 & Value        \\
        \hline
        \multicolumn{2}{c}{\textbf{ES}}          \\
        \hline
        Policy network            & 2x128 - ReLU \\
        Distribution $\sigma$     & 10           \\
        Population $\lambda$      & 100          \\
        Parents $\mu$             & 50           \\

        \hline
        \multicolumn{2}{c}{\textbf{TD3}}         \\
        \hline
        Critic network            & 2x128 - ReLU \\
        Actor lr                  & 0.001        \\
        Critic lr                 & 0.0003       \\
        Training steps / gen      & 1000         \\
        Discount $\gamma$         & 0.99         \\
        Soft update $\tau$        & 0.005        \\
        TD3 policy noise $\sigma$ & 0.2          \\
        Replay buffer size        & 1000000      \\
        Batch size                & 256          \\
    \end{tabular}
    \caption{Hyperparameters.}
    \label{table:hyperparameters}
    \end{subtable}
    \hfill
    \begin{subtable}{0.45\linewidth}
        \centering
    \begin{tabular}{l|c|c}
        Env            & \gdr{} $\epsilon$ & \sgdr{} $\epsilon$ \\
        \hline
        \halfcheetah{} & 0.01              & 0.001              \\
        \walker{}      & 0.01              & 0.01               \\
        \hopper{}      & 0.1               & 0.01               \\
        \swimmer{}     & 0.001             & 0.001              \\
        \humanoid{}    & 0.001             & 0.001              \\
        \ant{}         & 0.01              & 0.001              \\
    \end{tabular}
    \caption{$\epsilon$ values used for \gdr{} and \sgdr{}.}
    \label{table:epsilon_hp}
    \end{subtable}
\end{table}

\subsection{Parallel TD3 algorithm}
\label{app:td3_code}

\begin{algorithm}[h]
\caption{Parallel TD3}
\label{alg:parallel_td3}

\KwData{Environment $\mathcal{E}$, Replay buffer $R$, Actor $a_\theta$ and critic $q_\phi$, Number of generations $G$, Population size $pop\_size$, Batch size $B$, Number of steps $N\_steps$, Exploration noise $\epsilon$}
\BlankLine

\For{$g \leftarrow 1$ \KwTo $G$}{
    \For{$r \leftarrow 1$ \KwTo $pop\_size$}{
        Rollout of $a_\theta$ in $\mathcal{E}$ with exploration noise, store transitions in $R$\;
    }
    \For{$i \leftarrow 1$ \KwTo $N\_steps$}{
        Sample $B$ transitions from $R$\;
        Update $\phi$ with a gradient step on the critic loss\;
        Update $\theta$ with a gradient step on the actor loss\;
    }
}

\end{algorithm}

\subsection{Additional results}
\label{app:results}

\begin{table*}
\centering
\begin{tabular}{l|cccccc}
 & \halfcheetah{} & \walker{} & \hopper{} & \swimmer{} & \humanoid{} & \ant{}\\
\hline
GDR $\epsilon$=0.1 & 3975 (919) & 2549 (519) & 1434 (310) & 108 (0) & 591 (112) & -7 (3)\\
GDR $\epsilon$=0.01 & \textbf{4296} (286) & 2642 (353) & 1219 (387) & 109 (0) & 670 & \textbf{22} (75)\\
GDR $\epsilon$=0.001 & 3699 (950) & 2152 (155) & 1385 (189) & \textbf{109} (0) & \textbf{741} (83) & -10 (4)\\
GDR² $\epsilon$=0.1 & 4094 (476) & 2589 (450) & \underline{1494} (266) & 109 (0) & 646 (151) & -8 (3)\\
GDR² $\epsilon$=0.01 & 3739 (1271) & \textbf{2679} (440) & \textbf{1511} (291) & 108 (0) & 690 (43) & -11 (2)\\
GDR² $\epsilon$=0.001 & \underline{4175} (457) & \underline{2643} (523) & 1341 (334) & \underline{109} (0) & \underline{708} (160) & \underline{-4} (0)\\
\end{tabular}
\caption{Average return of the final policy found by each algorithm on each environment. Format: mean (std), best score in bold and second best underlined.}
\label{tab:results}
\end{table*}

We gave final fitness values for \gdr{}, \sgdr{} and baselines on \ant{}, \swimmer{} and \humanoid{} in the main paper. Learning curves for these environments are shown in Figure \ref{fig:app_results}. We also plot the weight of the actor in each update, which is based on its fitness. A non-zero weigh means it belonged to the best half of the population. \canonical{} uses exponentially decreasing weights similar to \cmaes{}, with no negative weights. Actor weight plots are reported in Figure \ref{fig:actor_weight}.
Fitness scores for different values of $\epsilon$ for \gdr{} and \sgdr{} are detailed in Table \ref{table:epsilon} .

\def\figwidth{0.3\linewidth}
\def\xleg{Evaluations}
\def\yleg{Fitness}
\begin{table*}
    \centering
    \begin{tabular}{cccc}
                                                                                                       & \ant{} & \swimmer{} & \humanoid{} \\
        \raisebox{5\normalbaselineskip}[1cm][0cm]{\rotatebox[origin=c]{90}{\vspace{1cm}ES fitness}}    &
        \includegraphics[width=\figwidth]{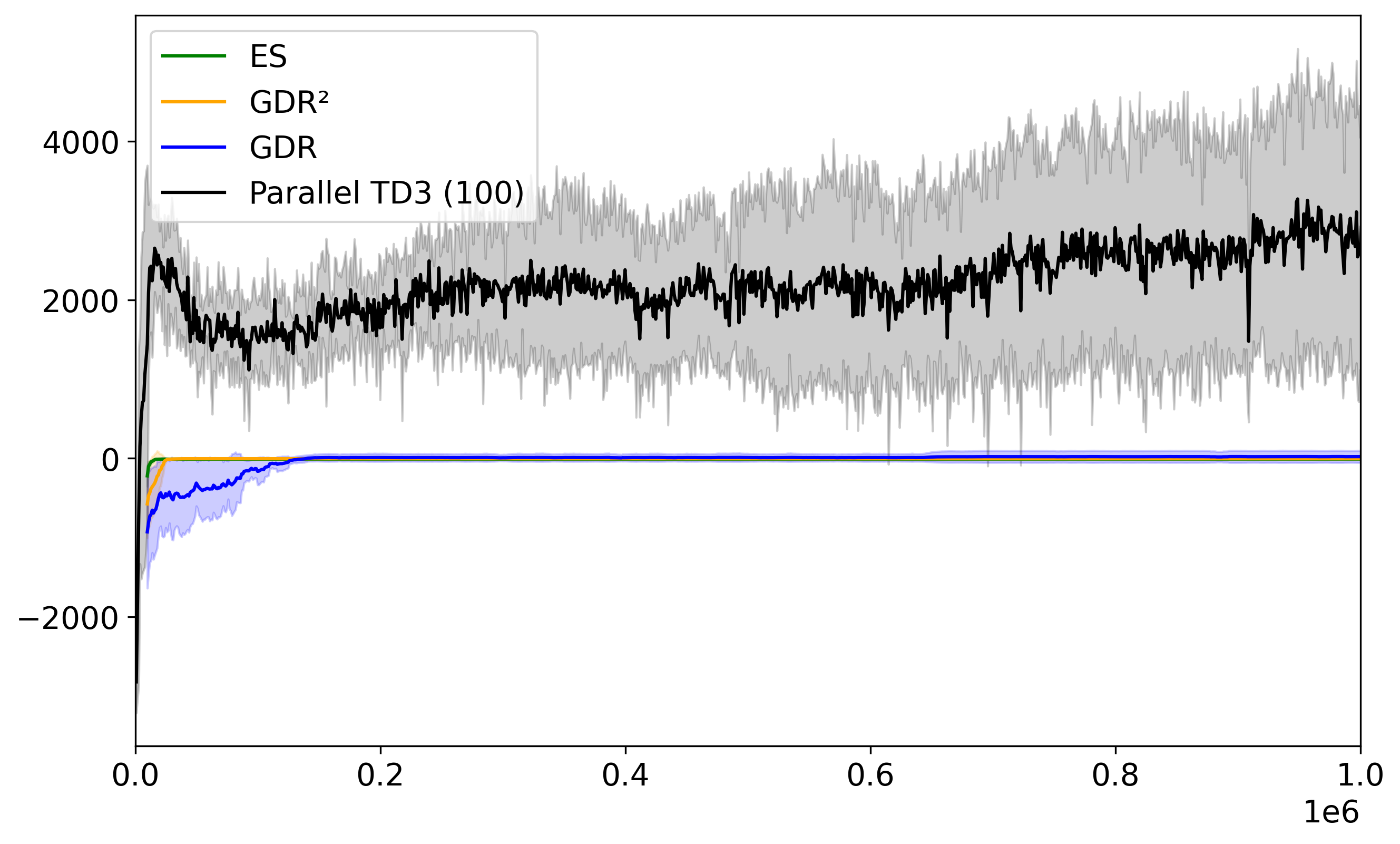}                                 &
        \includegraphics[width=\figwidth]{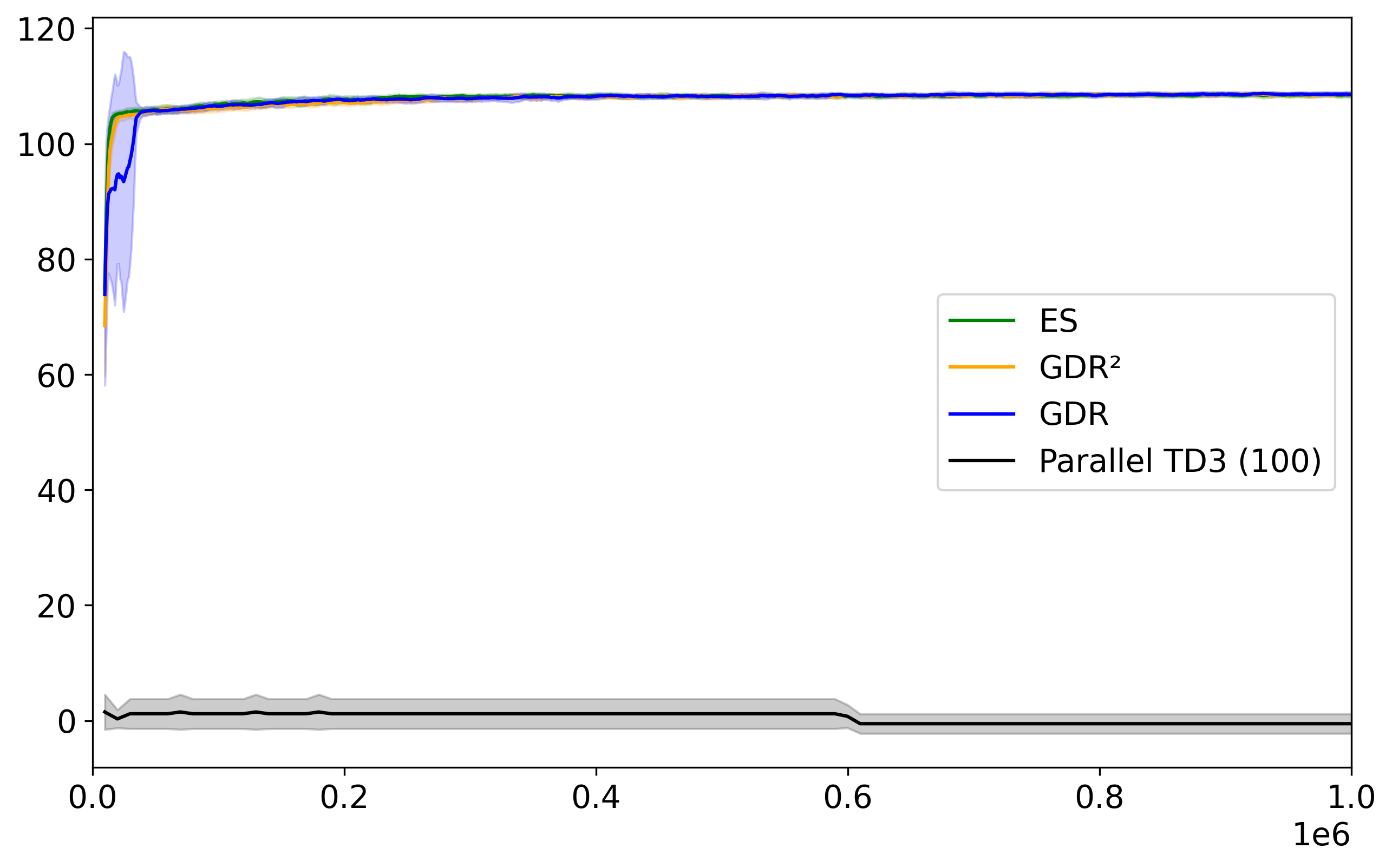}                              &
        \includegraphics[width=\figwidth]{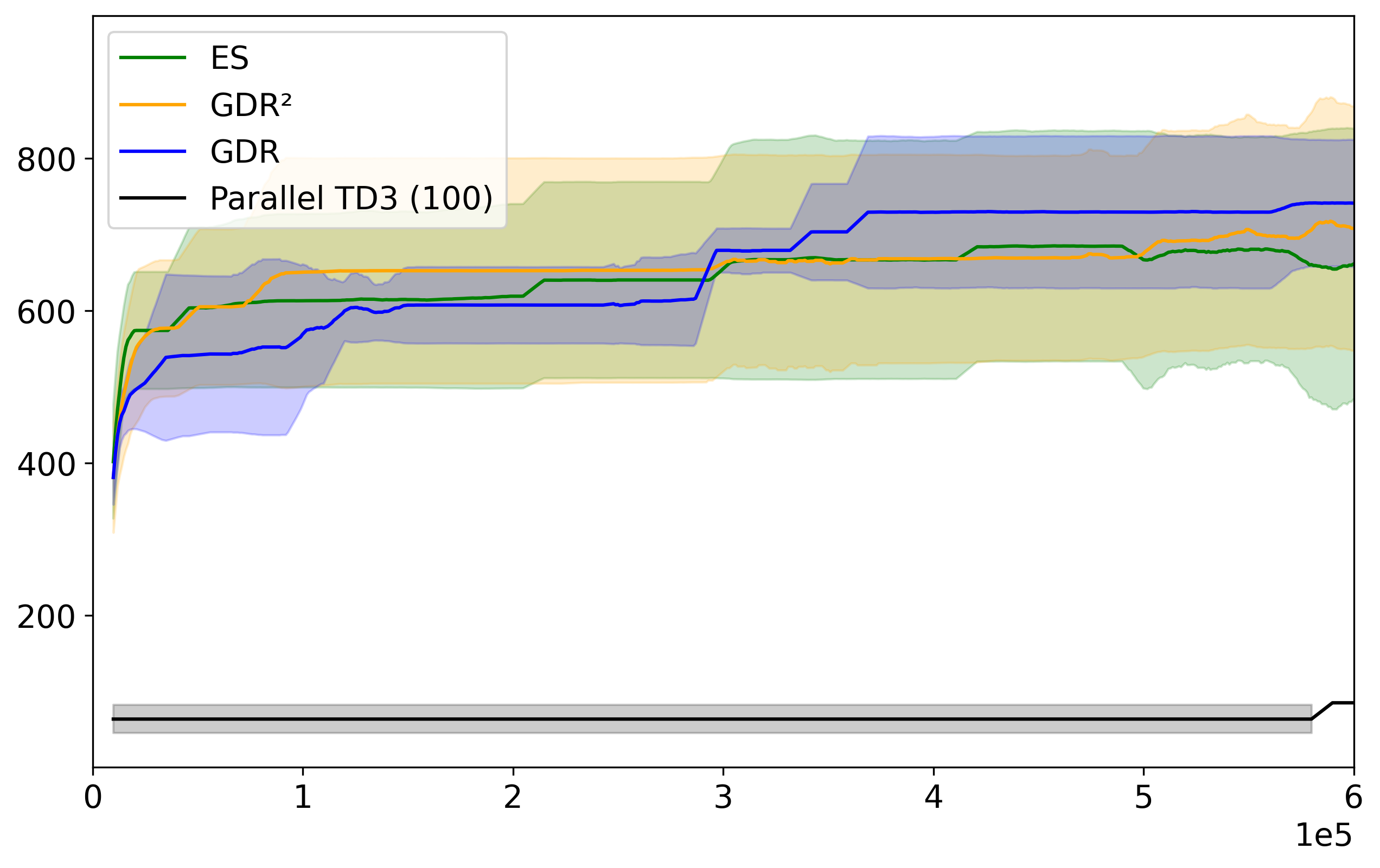}                                                                \\

        \raisebox{5\normalbaselineskip}[1cm][0cm]{\rotatebox[origin=c]{90}{\vspace{1cm}ES fitness}}    &
        \includegraphics[width=\figwidth]{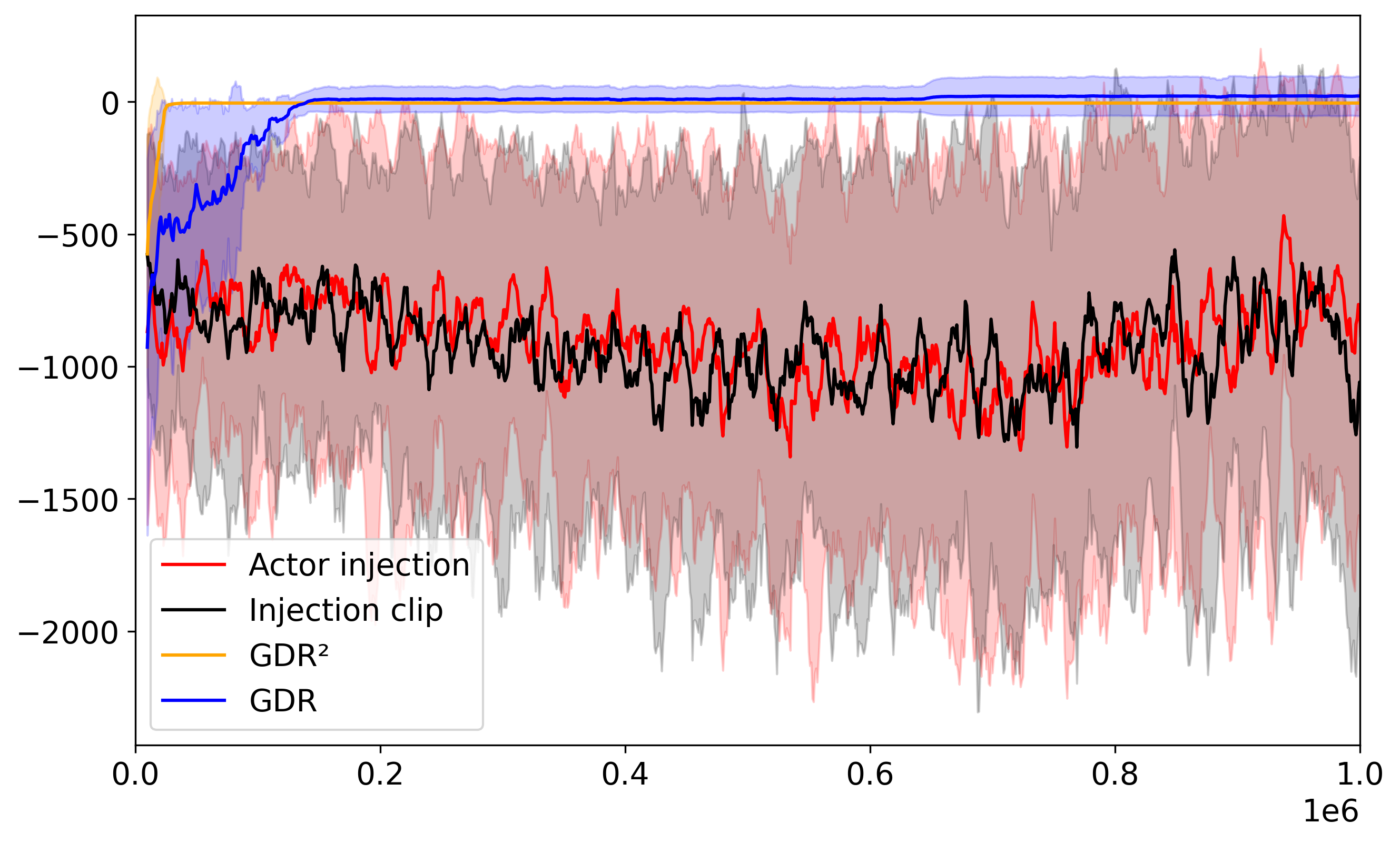}                                 &
        \includegraphics[width=\figwidth]{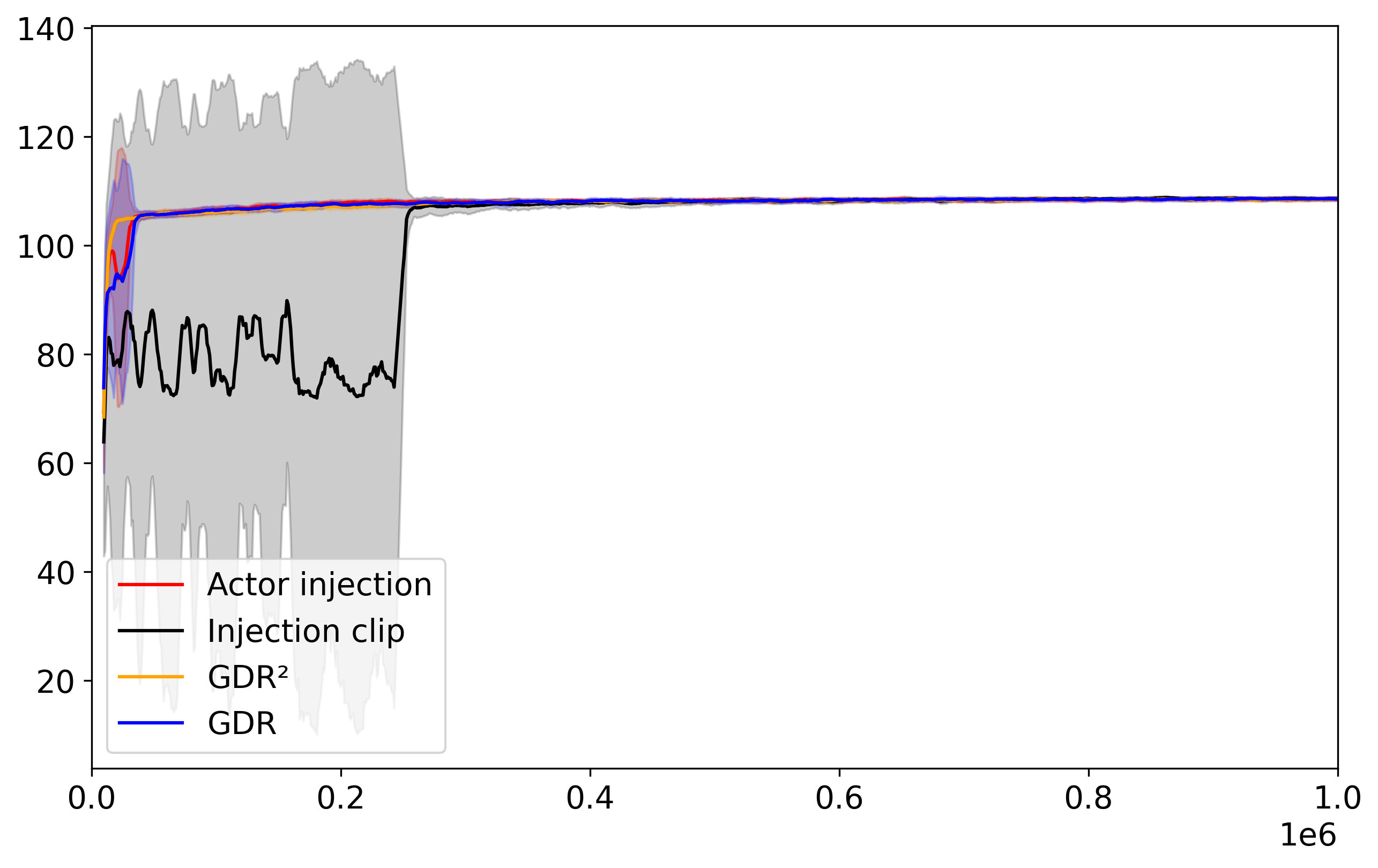}                              &
        \includegraphics[width=\figwidth]{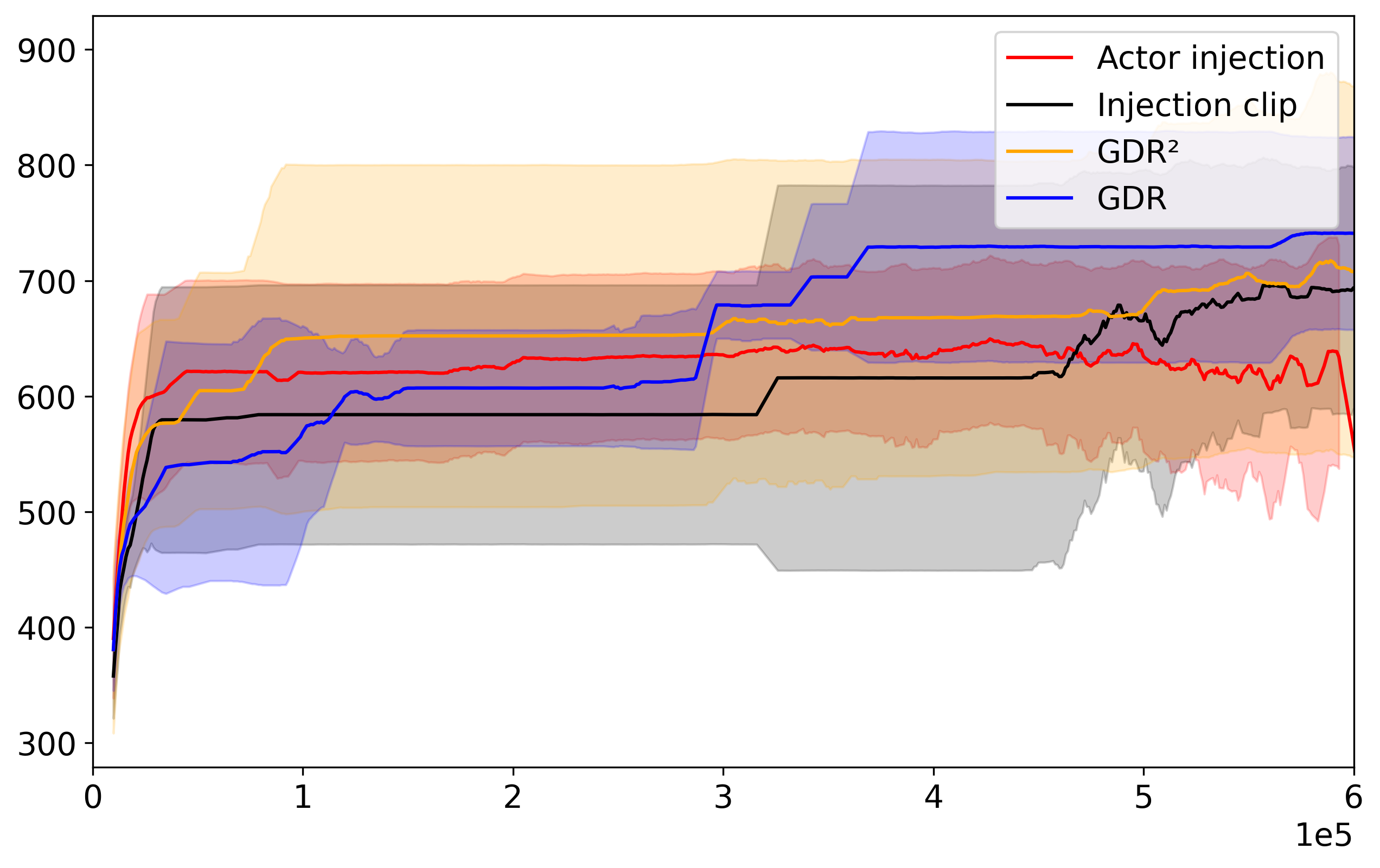}                                                                \\

        \raisebox{5\normalbaselineskip}[1cm][0cm]{\rotatebox[origin=c]{90}{\vspace{1cm}Actor fitness}} &
        \includegraphics[width=\figwidth]{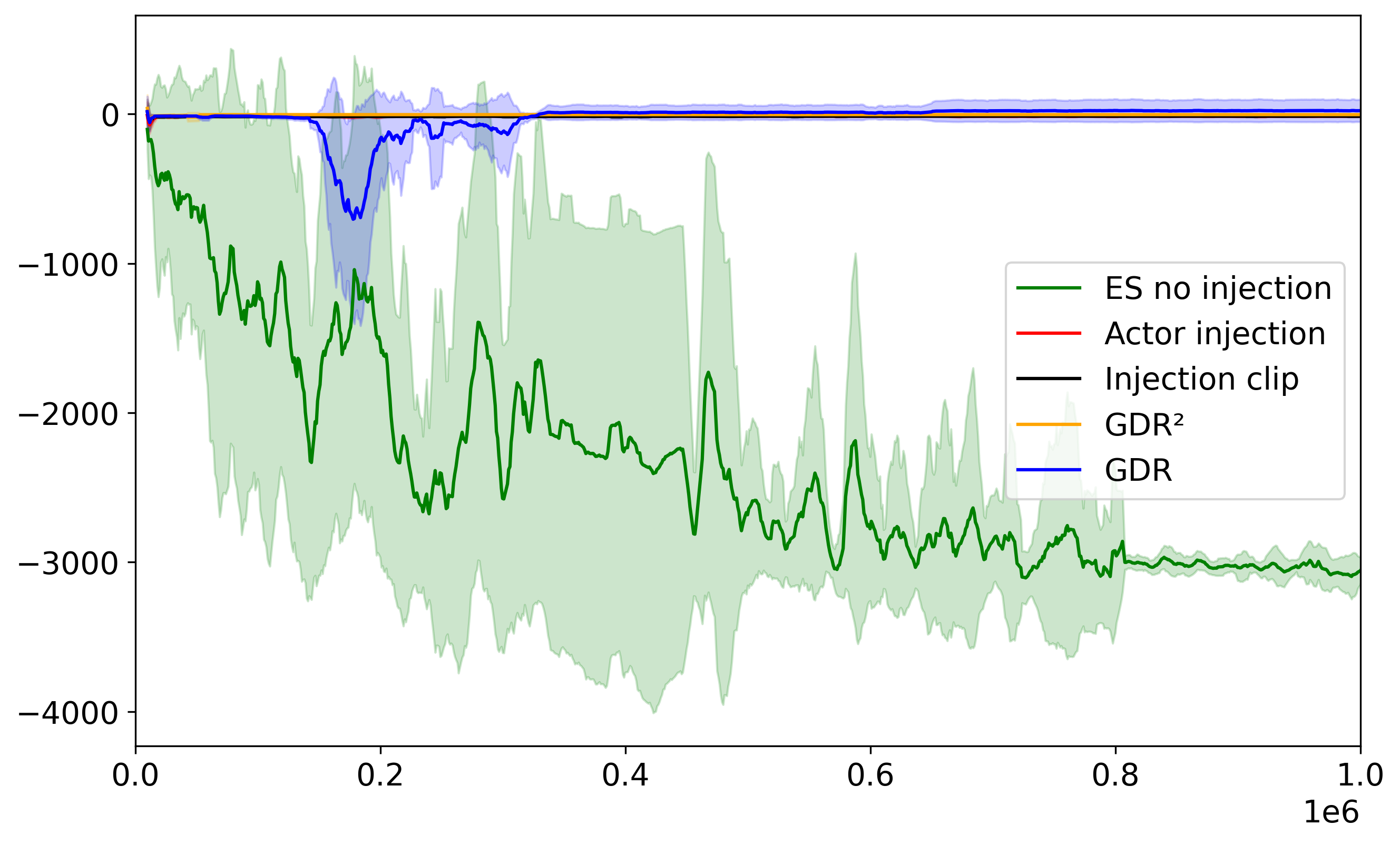}                                     &
        \includegraphics[width=\figwidth]{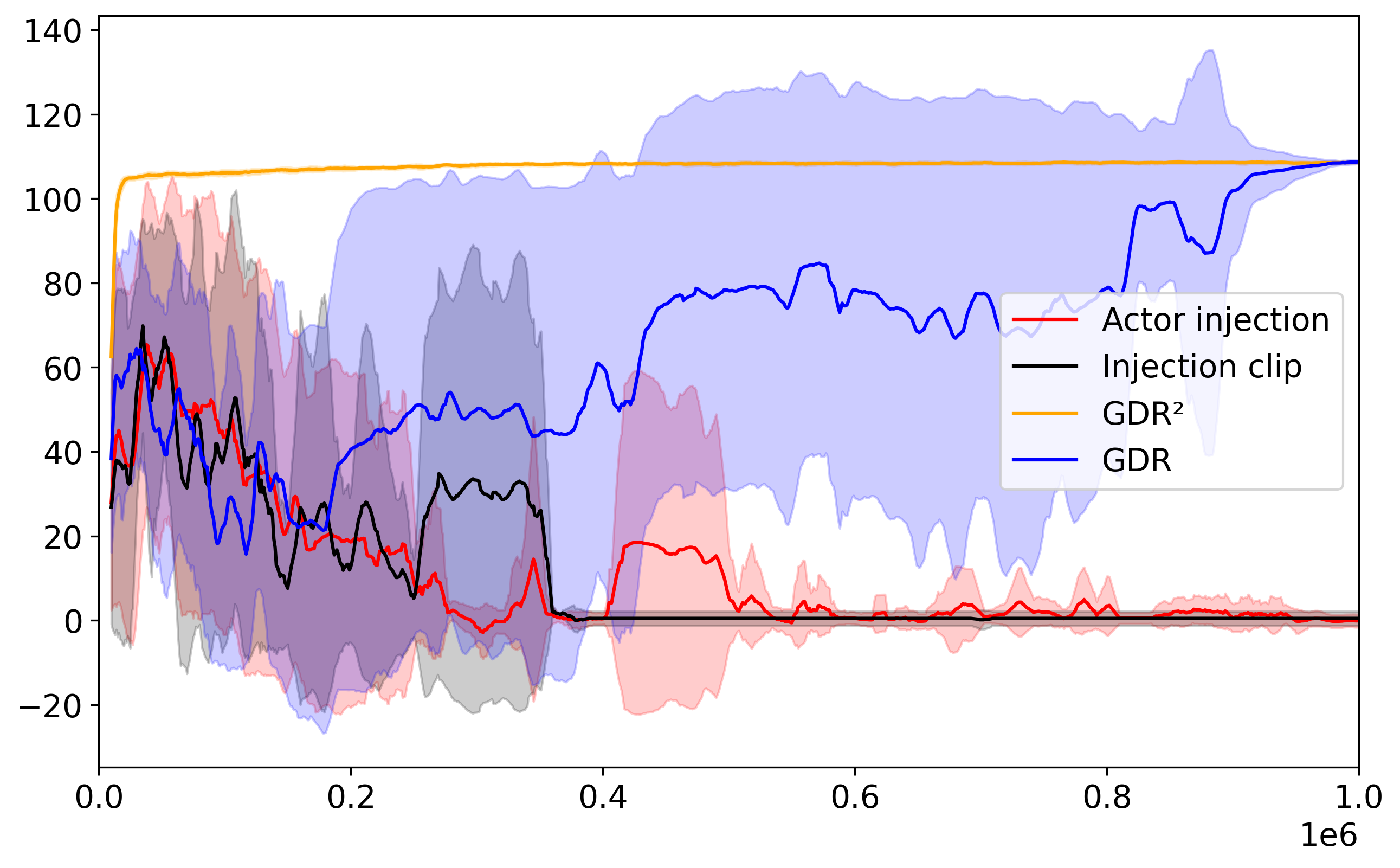}                                  &
        \includegraphics[width=\figwidth]{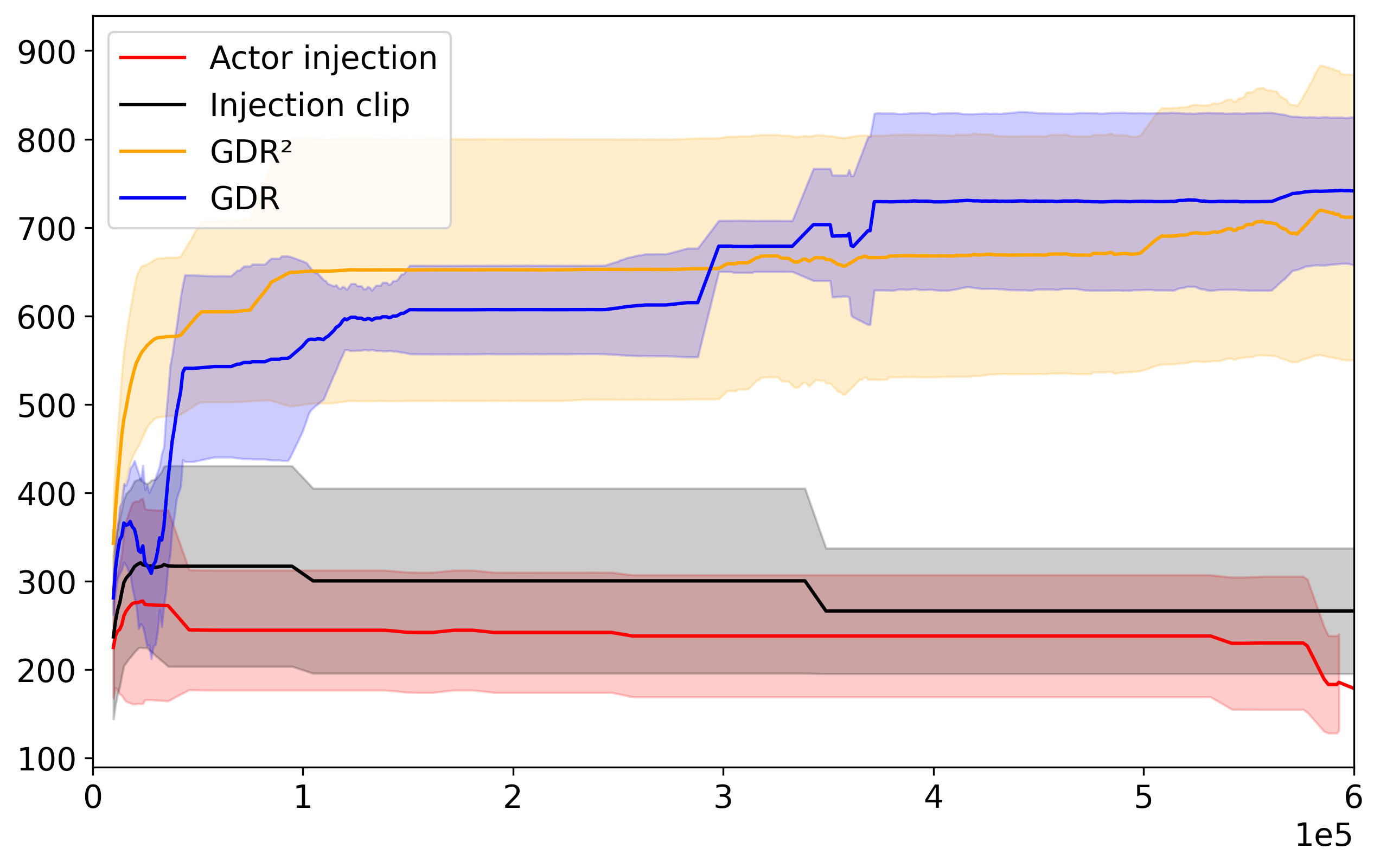}                                                                    \\
                                                                                                       & \xleg  & \xleg      & \xleg
    \end{tabular}
    \captionsetup{type=figure}
    \caption{Fitness of the ES center and the RL actor during training on 3 tasks. First and second rows compare \gdr{} and \sgdr{} to no-injection baselines and injection baselines respectively for readability. Third row shows the fitness of the actor during training. Colored areas show plus and minus one standard deviation.}
    \label{fig:app_results}
\end{table*}

\def\figwidth{0.3\linewidth}
\def\xleg{Evaluations}
\def\yleg{Weight}
\begin{table*}[h!]
    \centering
    \begin{tabular}{cccc}
                                                                                               & \halfcheetah{} & \walker{} & \hopper{} \\
        \raisebox{5\normalbaselineskip}[1cm][0cm]{\rotatebox[origin=c]{90}{\vspace{1cm}\yleg}} &
        \includegraphics[width=\figwidth]{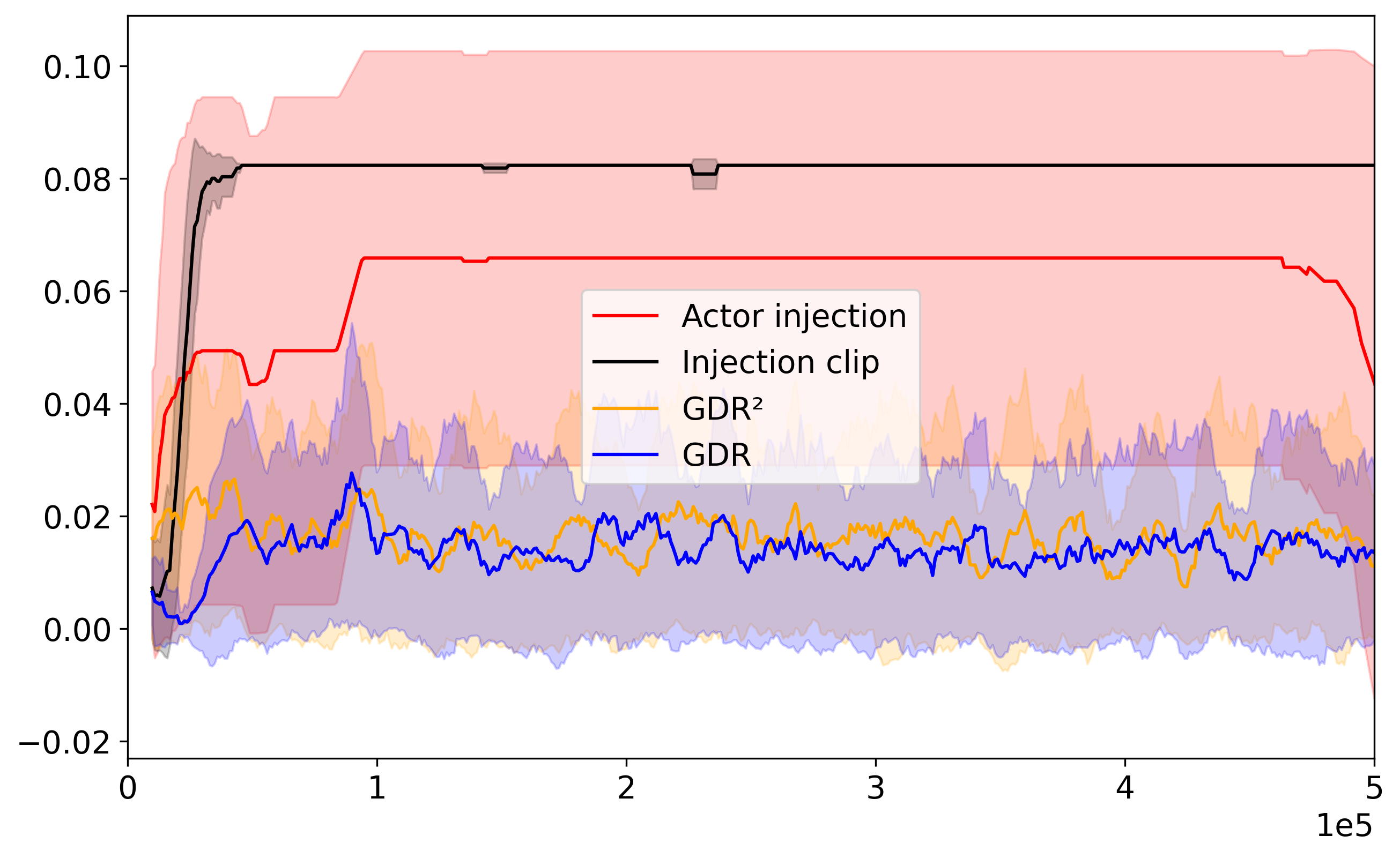}                    &
        \includegraphics[width=\figwidth]{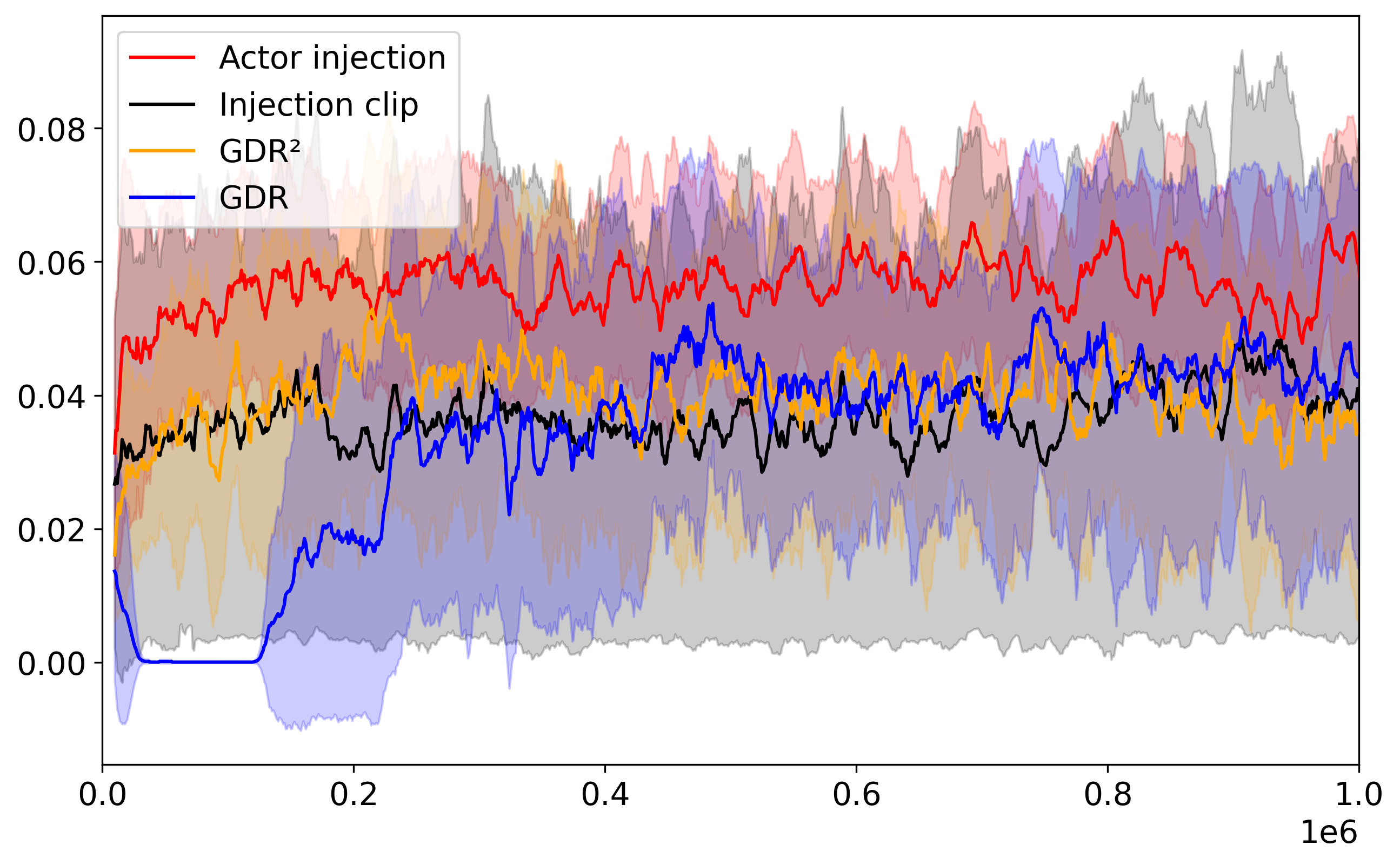}                       &
        \includegraphics[width=\figwidth]{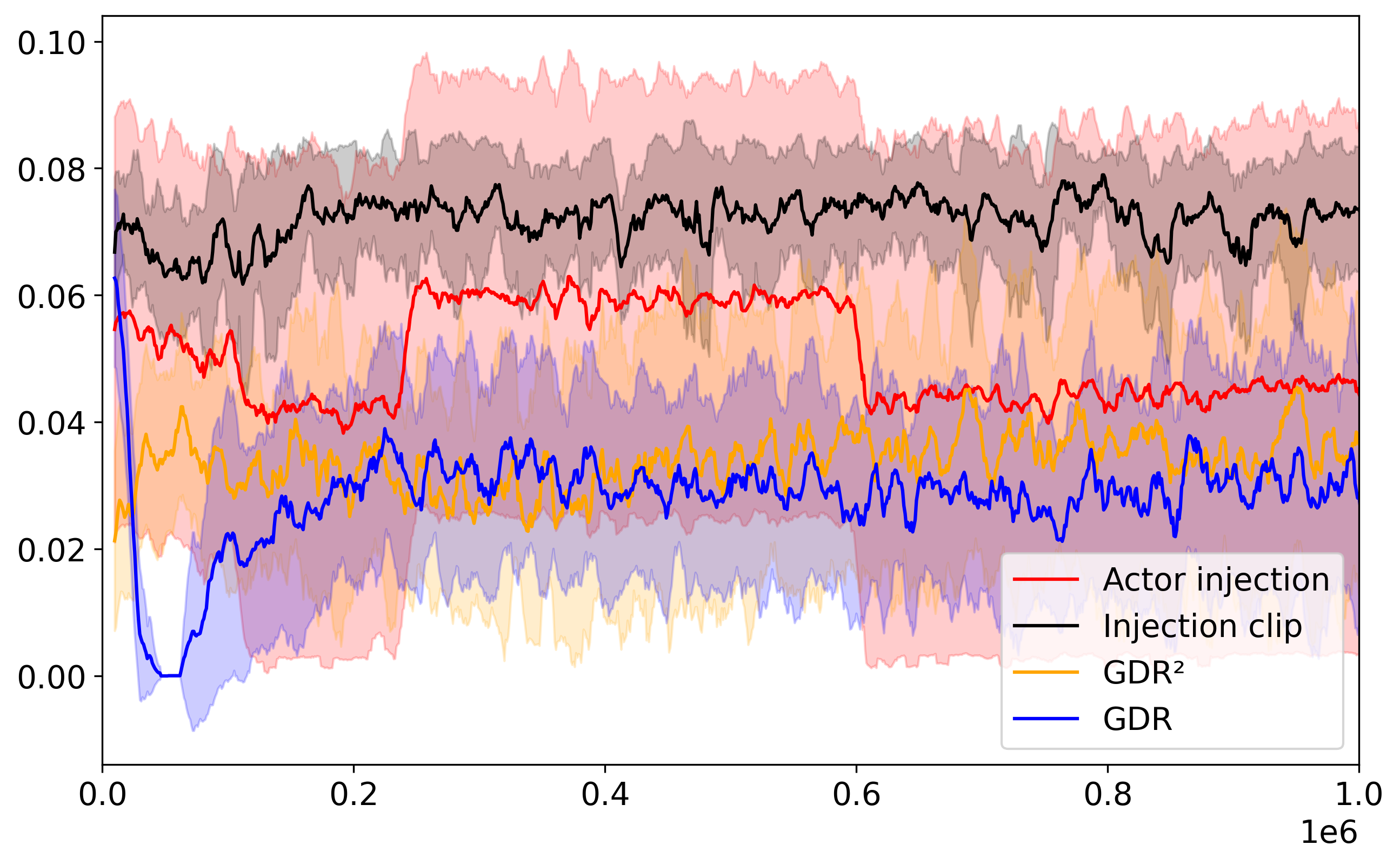}                                                                  \\
                                                                                               & \xleg          & \xleg     & \xleg
    \end{tabular}
    \captionsetup{type=figure}
    \caption{Weight of the RL actor in the ES update during training.}
    \label{fig:actor_weight}
\end{table*}

\end{document}